\documentclass[lettersize,journal]{IEEEtran}
\usepackage{amsmath,amsfonts}
\usepackage{array}
\usepackage{textcomp}
\usepackage{stfloats}
\usepackage{url}
\usepackage{verbatim}
\usepackage{graphicx}
\usepackage{cite}

\usepackage{graphicx}
\usepackage{amsmath}
\usepackage{amssymb}
\usepackage{booktabs}
\usepackage{subcaption}
\usepackage{multirow}
\usepackage[noend]{algpseudocode}
\usepackage{algorithmicx,algorithm}

\def\I{\mathbf{I}}
\def\N{\mathbf{N}}
\def\F{\mathbf{F}}
\def\K{\mathbf{K}}

\usepackage[capitalize]{cleveref}
\crefname{section}{Sec.}{Secs.}
\Crefname{section}{Section}{Sections}
\Crefname{table}{Table}{Tables}
\crefname{table}{Tab.}{Tabs.}

\begin{document}

\title{Learning Weighting Map for Bit-Depth Expansion within a Rational Range}

\author{
	Yuqing~Liu,
	Qi~Jia,
	Jian~Zhang,
	Xin Fan,~\IEEEmembership{Senior Member,~IEEE,}
	Shanshe Wang,
	Siwei~Ma,~\IEEEmembership{Member,~IEEE,}
	and~Wen~Gao,~\IEEEmembership{Fellow,~IEEE}
	\thanks{Y. Liu is with the School of Software, Dalian University of Technology,
		Dalian 116620, China (e-mail:liuyuqing@mail.dlut.edu.cn).}
	\thanks{Q. Jia and X. Fan are with International School of Information Science and Engineering, Dalian University of Technology, Dalian 116620, China (e-mail:
		jiaqi@dlut.edu.cn; xin.fan@dlut.edu.cn).}
	\thanks{J. Zhang is with the School of Electronic and Computer Engineering, Peking University Shenzhen Graduate School, Shenzhen, China. (e-mail: zhangjian.sz@pku.edu.cn)}
	\thanks{S. Wang, S. Ma, and W. Gao are with the School of Electronics Engineering and Computer Science, Institute of Digital Media, Peking University, Beijing 100871, China (e-mail: sswang@pku.edu.cn; swma@pku.edu.cn; wgao@pku.edu.cn).}

}

\markboth{Journal of \LaTeX\ Class Files,~Vol.~14, No.~8, August~2021}%
{Shell \MakeLowercase{\textit{et al.}}: A Sample Article Using IEEEtran.cls for IEEE Journals}

\IEEEpubid{0000--0000/00\$00.00~\copyright~2021 IEEE}

\maketitle

\begin{abstract}
		Bit-depth expansion (BDE) is one of the emerging technologies to display high bit-depth (HBD) image from low bit-depth (LBD) source. Existing BDE methods have no unified solution for various BDE situations, and directly learn a mapping for each pixel from LBD image to the desired value in HBD image, which may change the given high-order bits and lead to a huge deviation from the ground truth. In this paper, we design a bit restoration network (BRNet) to learn a weight for each pixel, which indicates the ratio of the replenished value within a rational range, invoking an accurate solution without modifying the given high-order bit information. To make the network adaptive for any bit-depth degradation, we investigate the issue in an optimization perspective and train the network under progressive training strategy for better performance. Moreover, we employ Wasserstein distance as a visual quality indicator to evaluate the difference of color distribution between restored image and the ground truth. Experimental results show our method can restore colorful images with fewer artifacts and false contours, and outperforms state-of-the-art methods with higher PSNR/SSIM results and lower Wasserstein distance. The source code will be made available at \url{https://github.com/yuqing-liu-dut/bit-depth-expansion}
\end{abstract}


\section{Introduction}
\begin{figure}[t]
	\centering
	\includegraphics[width=\linewidth]{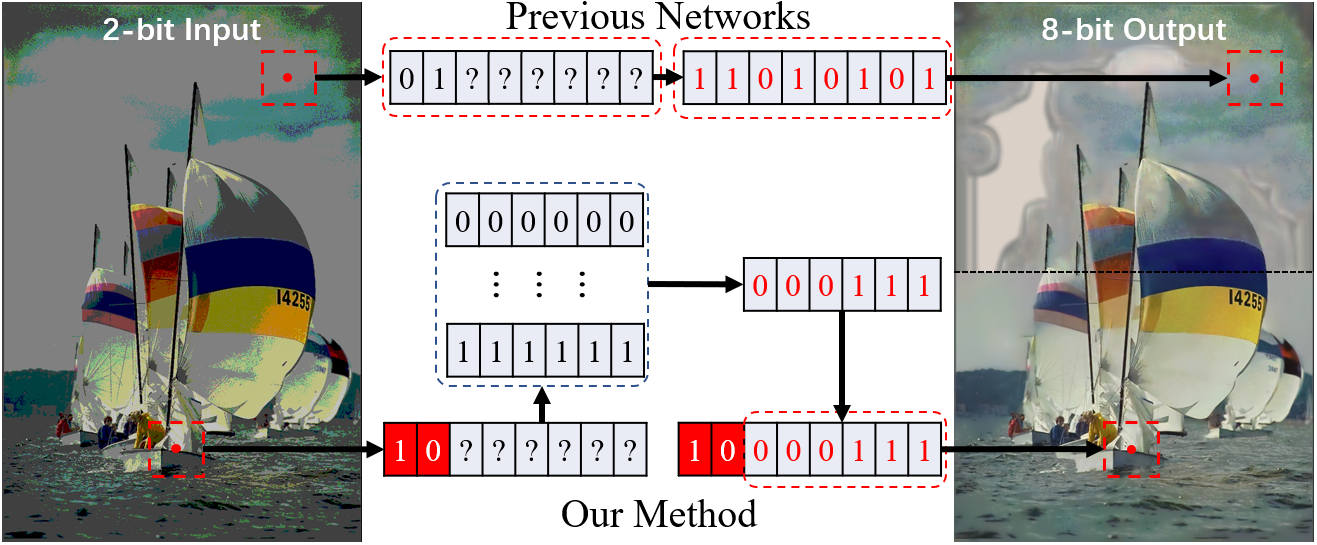}
	\caption{Comparison between our method and a representative work RMFNet~\cite{rmfnet_csvt2021} on restoring 2-bit image to 8-bit, showing our result on bottom part and the other on top part. RMFNet learns a direct mapping from low bit-depth to high bit-depth, that may change the given high-order bits, such as $01$ turns into $11$ in the instance. Our network learns a weight in a rational range and replenishes the missing bits without modifying the given information.}
	\label{fig:slogan}
\end{figure}

\IEEEPARstart{B}{it} depth expansion (BDE) is an emerging issue to generate and display a high bit-depth (HBD) image from the low bit-depth (LBD) source. With the rapid development of television technology, there are numerous ultra high-definition display terminals supporting 10-bit or even 12-bit luminance levels, while the mainstream video and image sources are just 8-bit~\cite{hdr_cvpr2020w}. Thus, expanding the bit depth becomes a popular topic for high-dynamic-range (HDR) display~\cite{ipad_tip2018}, display acceleration on mobile (SoC)~\cite{display_iscas2015} and video compression~\cite{ebda_icip2019, icassp09, tip11}. Given a LBD image, the task of BDE is to replenish the missing low-order bits according to the given high-order bits, and restore a suitable HBD representation~\cite{bden_tip2019}. \cref{fig:slogan} shows an example that restores an 8-bit HBD image from a 2-bit LBD instance. Existing methods still suffer from blur, artifacts and false contours caused by color quantization~\cite{lbden_tcsvt2021, bden_tip2019}.

Convolutional neural networks (CNNs) offer a fresh chance to break the deadlock of BDE~\cite{bden_tip2019, be_acgan_displays2021, vbde_tmm2019, tanet_tmm2021}, which directly learn a desired value for each pixel in the HBD image. For example, BitNet~\cite{bitnet_accv2018} considers the bit information as a prior to fit different bit-depth situations. LBDEN~\cite{lbden_tcsvt2021} utilizes a simple but effective network for BDE. However, these methods may modify the given high-order bits and lead to a huge deviation from the ground truth. \cref{fig:slogan} shows restoring a 2-bit image to 8-bit by a representative state-of-the-art CNN-based method RMFNet~\cite{rmfnet_csvt2021}. The result on top part demonstrates obvious banding effect and false contours. Before and after recovery, the bit value of a pixel instance indicates that the highest two bits are totally changed, demonstrating severe color distortions. 

\IEEEpubidadjcol

Moreover, most of the existing works have no adaptability on all bit-depth situations, and models are trained separately for specific bit-depth recovery. The degradation changes along with the number of missing low-order bits~\cite{bitplane_pami2021}, and previous works employ training different models for different bit-depth degradation~\cite{bden_tip2019, rmfnet_csvt2021}. Recently, Punnappurath and Brown train several independent networks to predict each missing bit for BDE~\cite{bitplane_pami2021}. However, to our best knowledge, no previous work finds an unified solver for all degradation situations, or designs a general-purpose network with good performance on all bit-depth degradation.

In this paper, we consider BDE in an information replenishing perspective. By keeping the correct high-order bits, we can restore more accurate value and generate the desired HBD image, as shown on bottom part of \cref{fig:slogan}. To find an unified representation for any bit-depth degradation, we learn the weight between the theoretical lower and upper bounds which is independent to the number of missing bits. Thus, the outputs from all bit depths hold an unified representation and we can use only one model to adaptively fit all degradation situations.

By considering BDE as an unified representation, it can be solved in the optimization perspective. However, the restoration task is usually a non-convex issue with a complex solution space and hard to find an optimal result. Thus, inspired by curriculum learning~\cite{curriculum_learning_icml2009}, we propose a progressive training strategy to gradually find a feasible solution with the increase of missing bits. Specifically, we design an end-to-end network for bit restoration (BRNet) and learn a weighting map for replenishing the LBD image to HBD image. Meanwhile, we devise an optimization block (OptBlock) based on the proximal gradient descent algorithm~\cite{prox_nips2020} and fourth-order Runge-Kutta (RK-4) method~\cite{rk4_cacm1966}. Besides, we introduce Wasserstein distance~\cite{wasserstein_aistas2021} to estimate the restored HBD image visual quality, which can reflect the pixel value distribution of the whole image other than pixel-wise evaluation as PSNR and SSIM. Experimental results show our network achieves better objective and subjective performances than state-of-the-art methods. The visual comparisons show our BRNet can efficiently reconstruct the details without false contour artifacts.

Our contributions are concluded as follows:

\begin{itemize}
	\item To our best knowledge, this weighting map learning network is the first to explore a general model to effectively restore any bit-depth image for BDE without modifying the given information. 
	
	\item We analyze the BDE problem in an optimization perspective and devise an optimization block (OptBlock) based on the proximal gradient descent algorithm and forth-order Runge-Kutta method, invoking an unified representation for any bit-depth degradation.
		
	\item By gradually increasing the number of missing bits, the tailored progressive training manner boosts the training phase and improves the network performance over full bit-depth recovery.
	
	\item Experimental results show our network achieves better objective and subjective performances than state-of-the-art methods with more satisfactory details and colors.
\end{itemize}

\begin{figure*}[t]
	\centering
	\includegraphics[width=.7\linewidth]{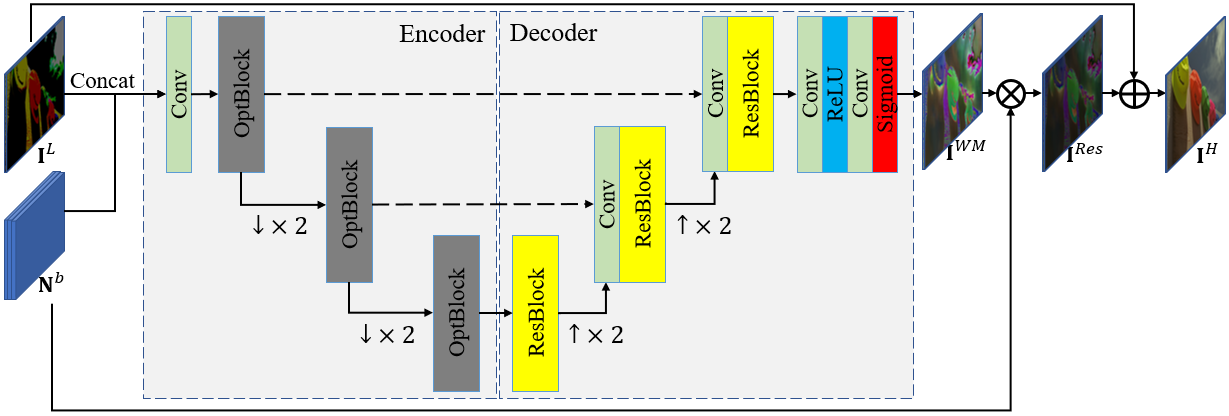}
	\caption{Network design of BRNet. An UNet-style architecture is designed to learn the weighting map for replenishing the LBD image to HBD.}
	\label{fig:net}
\end{figure*}

\begin{figure}[t]
	\centering
	\begin{subfigure}{0.2\linewidth}
		\centering		
		\includegraphics[width=\linewidth]{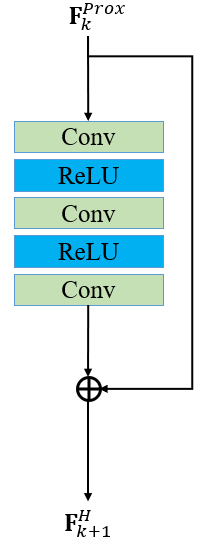}		
		\caption{ProxBlock}
		\label{fig:prox}
	\end{subfigure}
	\begin{subfigure}{0.47\linewidth}
		\centering				
		\includegraphics[width=\linewidth]{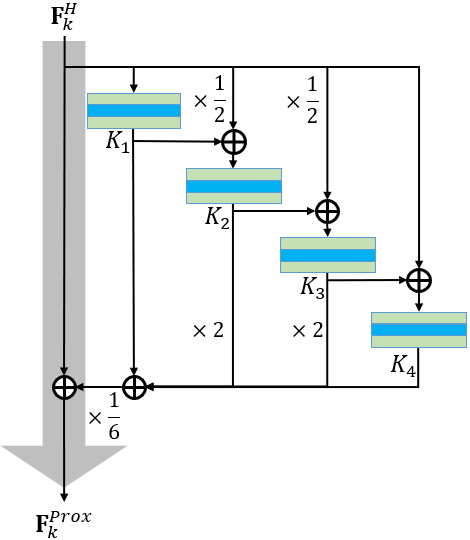}		
		\caption{RK-4 Block}
		\label{fig:rk4}
	\end{subfigure}
	\begin{subfigure}{0.28\linewidth}
		\centering		
		\includegraphics[width=\linewidth]{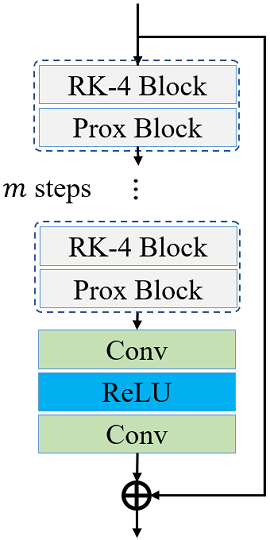}		
		\caption{OptBlock}
		\label{fig:optblock}
	\end{subfigure}
	\caption{Optimization-inspired block design. (a): Block for proximal operator (ProxBlock). (b): Block for RK-4 method (\text{RK-4 Block}). (c): OptBlock.}
	\label{fig:block}
\end{figure}

\section{Relate Works}
There are classical methods smoothing the banding effect of recovery image by filtering~\cite{bde_filtering_spl2021}, contour interpolation~\cite{bde_inter_icme2012}, and dithering~\cite{bde_dithering_tip2009}. Recently, IPAD~\cite{ipad_tip2018} proposed by Liu~\textit{et al.} introduces an intensity potential field to model the complicated relationships among pixels and utilizes adaptive de-quantization to convert LBD image to HBD, which achieves better performance than other traditional works. However, lacking of the learning and supervision process, there is still a big gap between the recovery image of classical methods and acceptable visual quality.

In recent years, CNN demonstrates great superiority on super-resolution~\cite{dpir_pami2021, cvpr_sr_1, cvpr_sr_9}, inpainting~\cite{plug_and_play_admm_tci2017, cvpr_ip_3, cvpr_ip_4}, denoising~\cite{deepam_cvpr2017} and other tasks~\cite{cvpr_ir_2, cvpr_ir_5}. Therefore, there are CNN-based BDE methods for effective restoration, which obtain the state-of-the-art performance. BitNet~\cite{bitnet_accv2018} uses an encoder-decoder architecture with dilated convolutions to reconstruct the HBD images GG-DCNN~\cite{ggdcnn_accv2018} utilizes an UNet-style design to restore image with very low bit-depth. To avoid the information loss, single-scale networks without down-sampling operation are considered for better performance. BE-CNN~\cite{becnn_nc2019} enjoys a residual connection design to suppress the artifacts. BE-CALF~\cite{be_calf_tip2019} uses densely connections to directly generate the high quality images. Besides the single-scale designs, RMFNet~\cite{rmfnet_csvt2021} introduces a multi-scale network with residual-guided information fusion and achieves the state-of-the-art performance. These works treat different areas of the input image equally and generate the HBD results or residual images straightforwardly. BDEN~\cite{bden_tip2019} considers the flat and non-flat areas separately, and uses a two-stream restoration network to restore the information. LBDEN~\cite{lbden_tcsvt2021} is proposed for lighter but efficient bit-depth expansion, which achieves competitive performance to the heavier works. Recently, BitPlane~\cite{bitplane_pami2021} separates the image into different bit planes, and uses several independent networks to predict the missing bits from different positions. Besides the methods that restores the missing information, there are also GAN-based networks for BDE, aiming to generate the satisfactory results, such as BDGAN~\cite{bdgan_icme2020} and TSGAN~\cite{bde_bmsb2020}.

BDE is widely considered in video compression to save the bit rate and enhance the visual experience~\cite{icassp09, tip11}. Zhang \textit{et al.} devise EBDA-CNN~\cite{ebda_icip2019} and integrate the network into HEVC codec. Ma \textit{et al.} also consider the GAN-based method for perceptual video compression~\cite{bdgan_icme2020}.

\section{Methodology}
In the following, we introduce the methodology for BDE in detail. Firstly we describe the pipeline of learning the weighting map for BDE. Then, the overall design of BRNet and component design of OptBlock are discussed in detail. Finally, we introduce the proposed progressive training strategy and the quality assessment.

\subsection{Problem Formulation}

Let $\I^L$, $\I^H$ be the LBD and restored HBD images, respectively. Our method aims to find the missing information from lost bits and keep the given high-order bits. Let $\I^{Res}$ be the residual image of the replenished low-order bits, then there is
\begin{equation}
	\label{eq:1}
	\I^H = \I^L + \I^{Res}.
\end{equation}

The residual image is gained based on the learned weighting map $\I^{WM}$. Let $\N^b$ be the theoretical upper bound of the missing bits. Then, $\N^b$ is a 3-channel tensor denoting the RGB color channels separately. The value of $\N^b$ is the decimal representation of the missing low-order bits. For example, if the number of missing bits $b=4$, each value of $\N^b$ is $2^4=16$. Thus, the residual image $\I^{Res}$ is
\begin{equation}
	\label{eq:2}
	\I^{Res} = \N^b * \I^{WM},
\end{equation}
where $*$ denotes the point-wise multiplication, and $I^{WM}$ is the weighting map with each value in the range of $[0,1]$.

Further, we design a bit restoration network (BRNet), jointly considering $\I^L$ and the upper bound $\N^b$ to infer the adaptive weighting map $I^{WM}$. Let $f^{BRNet}(\cdot)$ represent the network, we can formulate the learning process as
\begin{equation}
	\label{eq:3}
	\I^{WM} = f^{BRNet}([\I^L, \N^b]),
\end{equation}
where $[\cdot]$ is the concatenation operation. 

\subsection{Network Design}
\cref{fig:net} illustrates the entire framework of the proposed BRNet. BRNet is mainly composed of encoder and decoder that follows an UNet-style design~\cite{unet_miccai2015}. The encoder contains an input convolutional layer and three optimization blocks. The input layer converts the input tensor to a series of feature maps for optimization. The optimization blocks (OptBlock) explore the feature distribution in different scale and find a feasible solution in the feature space. The optimized feature maps are down-scaled successively by MaxPooling and explored by the next stage. The smaller size of feature equivalently increases the receptive field of network and learns different scales of feature distributions. The decoder part integrates the multi-scale solutions and produces the desired weighting map.

The design of optimization block (OptBlock) follows the mathematical analysis of BDE in the optimization perspective. Let $\F^L$ be the LBD feature, the task of OptBlock is to find a restored HBD feature $\F^H$ satisfying
\begin{equation}
	\label{eq:4}
	\F^H = \arg\min_{\F^H} \frac{1}{2}||\mathcal{D}_b(\F^H)-\F^L||^2_\ell+\lambda\Phi(\F^H),
\end{equation}
where $\mathcal{D}_b(\cdot)$ is the operation that removes the information of last $b$ bits, $\Phi(\cdot)$ is the prior term, $\lambda$ is a weighting factor. It is worth noting that \cref{eq:4} is an unconstrained optimization issue, and the solution space of $\F^H$ has no extra boundary. By converting the feature into weighting map, we can find a suitable mapping within the rational range.

An ideal prior $\Phi(\cdot)$ should be a function related to the number of bits in $\F^H$, which can be regarded as a statistical-relative indicator. However, it is difficult for derivation. Thus, we employ proximal gradient descent algorithm to solve \cref{eq:4} in an iterative formulation by
\begin{equation}
	\label{eq:5}
	\F^H_{k+1}=\mathop{Prox}_{t_k, \lambda\Phi}(\F^H_k-t_k\nabla g(\F^H_k)),
\end{equation}
where
\begin{equation}
	\label{eq:6}
	\mathop{Prox}_{t_k, \lambda\Phi}(\mathbf{z_1}) = \arg\min_\mathbf{z_2} \frac{1}{2t_k} ||\mathbf{z_1}-\mathbf{z_2}||^2 + \lambda\Phi(\mathbf{z_2}),
\end{equation}
and
\begin{equation}
	\label{eq:7}
	g(\mathbf{z_3})=\frac{1}{2}||\mathcal{D}_b(\mathbf{z_3})-\F^L||_\ell^2.
\end{equation}
$\mathbf{z_1}$, $\mathbf{z_2}$, and $\mathbf{z_3}$ are placeholders.

\cref{eq:6} has a similar formulation to traditional image restoration problem, and thereby we design a residual block ProxBlock composed of three convolutional layers and two ReLU activation to find the solution, as demonstrated in \cref{fig:prox}.

Further, $\mathcal{D}_b(\cdot)$ can be regarded as a quantization step to remove the information of last $b$ bits in the feature space, which is difficult to find an explicit quantization expression. To address this issue, we denote $\F^{Prox}_k=\F^H_k-t_k\nabla g(\F^H_k)$, $h_k=-t_k$, and $G(\mathbf{u}, \mathbf{v})=\nabla g(\mathbf{v})$, then there is
\begin{equation}
	\label{eq:8}
	\F^{Prox}_k = \F^H_k + h_kG(\mathbf{u}_k, \F^H_k),
\end{equation}
where $\mathbf{u}$, $\mathbf{v}$ are placeholders.

\cref{eq:8} holds a similar formulation to the Euler method for finding an approximation of ordinary differential equation (ODE). Thus, we consider \cref{eq:8} as a dynamical system, and find the solution by a one-step approximation over an interval $h_k$. To find a more accurate numerical solution, we utilize fourth-order Runge-Kutta (RK-4) method~\cite{rk4_cacm1966} to calculate the result and there is
\begin{equation}
	\label{eq:9}
	\F^{Prox}_k=\F^{H}_k+\frac{h_k}{6}(\K_1+2\K_2+2\K_3+\K_4),
\end{equation}
where
\begin{equation}
	\label{eq:10}
	\left\{
	\begin{aligned}
		\K_1 &= G(\mathbf{u}_k, \F^H_k), \\
		\K_2 &= G(\mathbf{u}_k + \frac{h_k}{2}, \F^H_k + \frac{h_k}{2}\K_1), \\
		\K_3 &= G(\mathbf{u}_k + \frac{h_k}{2}, \F^H_k + \frac{h_k}{2}\K_2), \\
		\K_4 &= G(\mathbf{u}_k + h_k, \F^H_k+h_k\K_3). \\
	\end{aligned}
	\right.
\end{equation}

\cref{fig:rk4} shows the design for finding the solution by RK-4 (RK-4 Block). We emulate the $G(\cdot)$ of \cref{eq:10} by the color blocks of \cref{fig:rk4} where each block consists of two convolutional layers and a ReLU activation.

As shown in \cref{fig:optblock}, each optimization step consists of a group of one RK-4 Block and one ProxBlock, that optimizes the feature map according to \cref{eq:9} and \cref{eq:5}, separately. After $m$ optimization steps in the OptBlock, we design a residual block with two convolutional layers and a ReLU activation for better gradient transmission. 

There are also three stages in the decoder. For each decoder stage, a $1\times1$ convolutional layer and a residual block (ResBlock) integrate the information from the current stage of the encoder and the former decoder stage. The ResBlock follows the same design as EDSR~\cite{edsr_cvprw2017} which is composed of two convolutional layers and a ReLU activation. The up-scaling method in the decoder is devised by a $1\times1$ convolutional layers and a sub-pixel convolution~\cite{espcn_cvpr2016}. The $1\times1$ convolution expands the number of channels and the sub-pixel convolution reshapes the feature maps and increases the resolution. At the end of the decoder, there is a block group to restore the weighting map from the feature. Two convolutional layers, a ReLU activation and a Sigmoid activation are utilized to explore the decoded feature maps and generate the weighting map $\I^{WM}$.

In this paper, the filter number is set as 64 for all convolutional layers, except for the $1\times1$ convolutions and the last convolutional layer. For the first and second stages of the encoder, the number of optimization step is set as 1. For the third stage, there are 6 optimization steps in the OptBlock. The BRNet is designed for bit-depth expansion on standard dynamic range (SDR) image with 8 bits. For HDR bit-depth expansion (with 16 bits),  we use two parallel block groups at the end of the decoder for restoration, since it has double information than SDR. Similar to SDR, each group aims to restore at most 8-bit information. 

\subsection{Progressive Training Strategy}
It is obvious that replenishing a small number of bits is easier than restoring many of bits. Inspired by the human's learning behavior from easy to hard, we propose a progressive training strategy by gradually increasing the missing bits of the training data to boost the learning step~\cite{curriculum_learning_icml2009}. Let $\mathcal{S}=\{\mathcal{X}_1,...\mathcal{X}_P\}$ be $P$ datasets with different difficulty. The difficulty rises with the increase of the indicator $P$. Then, the training steps are regarded as a sequence $\mathcal{C}=<\mathcal{T}_1,...,\mathcal{T}_P>$, where $\mathcal{T}_P=(\theta, \mathcal{X}_P)$, meaning to update the set of network parameter $\theta$ with dataset $\mathcal{X}_P$. The indicator $P$ of training steps is sequentially increasing, and the set of network parameters $\theta$ is progressively optimized. Specially, the difficulty of $\mathcal{X}_P$ is defined as the upper bound of missing bits. With the increase of missing bits, the information suffers severer loss, and the restoration becomes harder.

The detail of progressive training is as follows. For $E$-th epoch, we generate the LBD image $\I^{L}_E$ from the original image $\I^{Ori}_E$ for training. Specifically, we increase the difficulty of training data for every 20 epochs. Let $b^{max}$ be the maximum bit depth of image, the upper bound of missing bits for $E$-th epoch $b^{ub}_E$ is calculated as $b^{ub}_E=\min(4+\lfloor E/20 \rfloor, b^{max}-1)$. $b^{max}$ is set as 8 for SDR image, and 16 for HDR image. Then, we randomly remove $b_E^{pt}$ bits for progressive training from $\I^{Ori}_E$ and generate the LBD image $\I^{L}_E=\mathcal{D}_{b^{pt}_E}(\I^{Ori}_E)$, where $b_E^{pt}={\rm rand}(1, b^{ub}_E)$. The overall progressive learning procedure is outlined in \cref{algo:cl}. Overall, we gradually increase the training difficulty to simulate the human's learning behavior for better performance.

\begin{algorithm}[t]
	\caption{Data Generation for Progressive Training} 
	\label{algo:cl}
	{\bf Input:} 
	The current epoch $E$, the maximum image bit depth $b^{max}$, and the original HBD image $\I^{Ori}_E$ for $E$-th epoch.\\
	{\bf Output:} 
	The degraded LBD image $\I^{L}_E$ for training.
	\begin{algorithmic}[1]
		\State Calculate the upper bound of missing bits for $E$-th epoch $b^{ub}_E=\min(4+\lfloor E/20 \rfloor, b^{max}-1)$.
		\State Randomly select the number of missing bits for $E$-th epoch for progressive training $b^{pt}_E={\rm rand}(1, b^{ub}_E)$.
		\State Generate the LBD image $\I^L_E=\mathcal{D}_{b^{pt}_E}(\I^{Ori}_E)$.
		\State \Return $\I^L_E$
	\end{algorithmic}
\end{algorithm}

\begin{figure}[t]
	\centering
	\includegraphics[width=.7\linewidth]{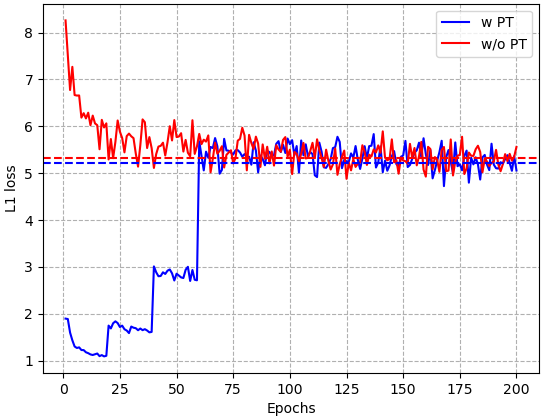}
	\caption{Loss comparison on progressive training (PT). The dash line denotes the converged loss value.}
	\label{fig:cl}
\end{figure}

\subsection{Quality Assessment}
Besides the network design, how to evaluate the performance of restored HBD image is also a vital issue for BDE~\cite{iqa_tip2015}. By delving into the evaluation method of the human visual system, we hold the hypothesis that if the restored image has the color distribution closer to the natural image, it is more coincident with visual experience. However, the existing PSNR and SSIM are both pixel-wise evaluations~\cite{bde_icassp2016, bde_bmsb2020}, having no capability of reflecting the distribution of the whole image. Thus, a distribution distance metric is more suitable for visual performance evaluation.

We use Wasserstein distance (W-dis) to indicate the difference between restored image $\I^H$ and the ground truth. Wasserstein distance describes the physical transportation cost between two distributions. Similar to PSNR and SSIM, Wasserstein distance is a symmetric metric, which is more suitable for image quality assessment than KL-divergence and other asymmetrical methods. \cref{fig:wdis} shows comparisons between PSNR/SSIM and W-dis. In the first row of \cref{fig:wdis}, The PSNR of BE-CALF is 0.1 dB higher than RMFNet, but there are more artifacts from BE-CALF in the background of the parrot. RMFNet holds lower W-dis than BE-CALF and has fewer unnatural textures, indicating that W-dis consists with the image quality. 

\begin{figure}[t]
	\centering
	\begin{subfigure}{0.32\linewidth}		
		\includegraphics[width=\linewidth]{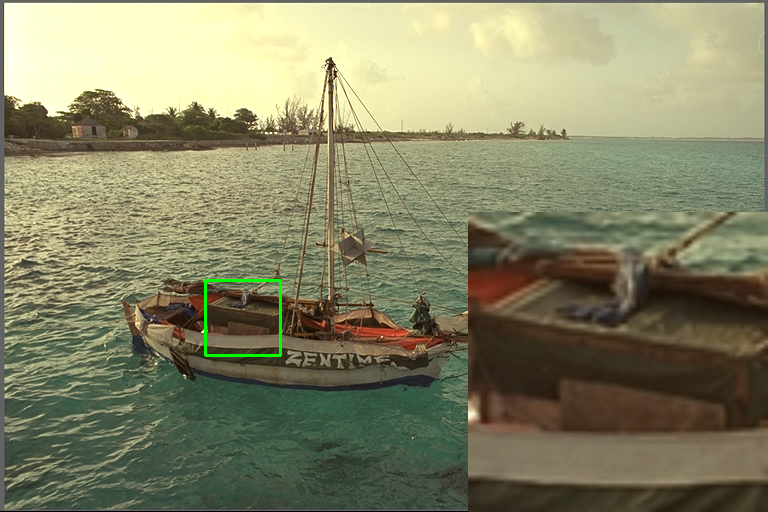}		
		\caption*{\centering{Original Image~\linebreak{(PSNR/W-dis)}}}
		\label{fig:5hbd}		
	\end{subfigure}
	\begin{subfigure}{0.32\linewidth}
		\centering				
		\includegraphics[width=\linewidth]{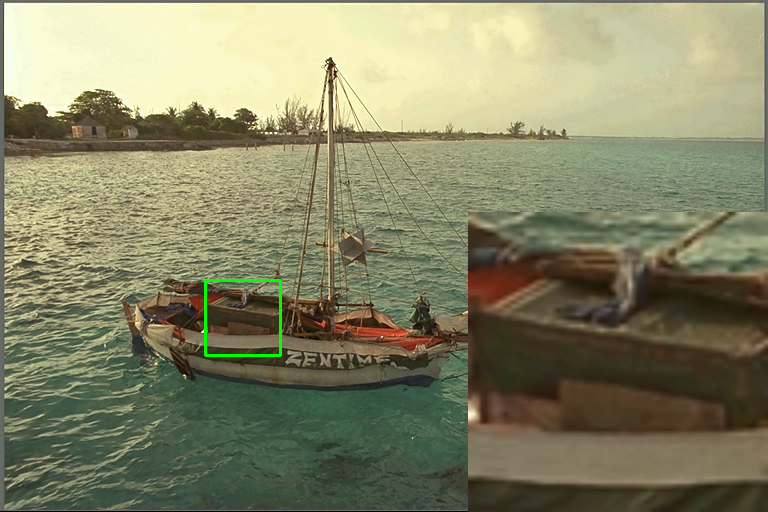}		
		\caption*{\centering{BE-CALF~\cite{be_calf_tip2019}~\linebreak{(38.29/1.40)}}}
		\label{fig:5lbd}
	\end{subfigure}
	\begin{subfigure}{0.32\linewidth}
		\centering		
		\includegraphics[width=\linewidth]{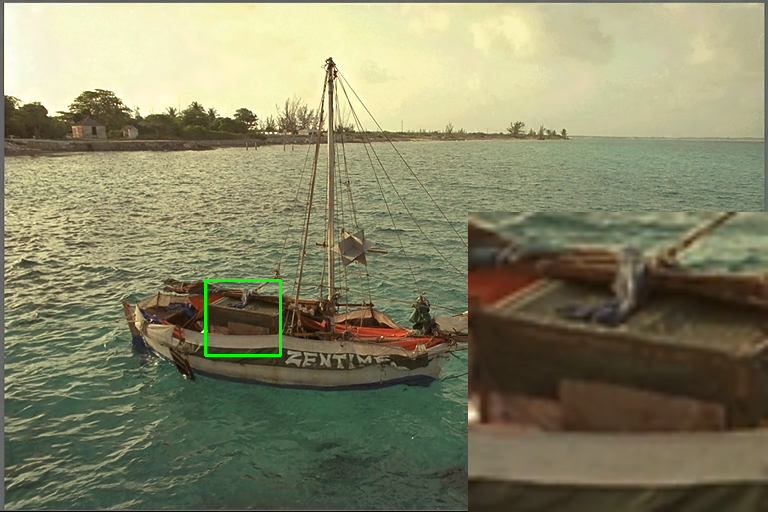}		
		\caption*{\centering{RMFNet~\cite{rmfnet_csvt2021}~\linebreak{(37.72/1.03)}}}
		\label{fig:5sbd}
	\end{subfigure}

	\begin{subfigure}{0.32\linewidth}
		\centering		
		\includegraphics[width=\linewidth]{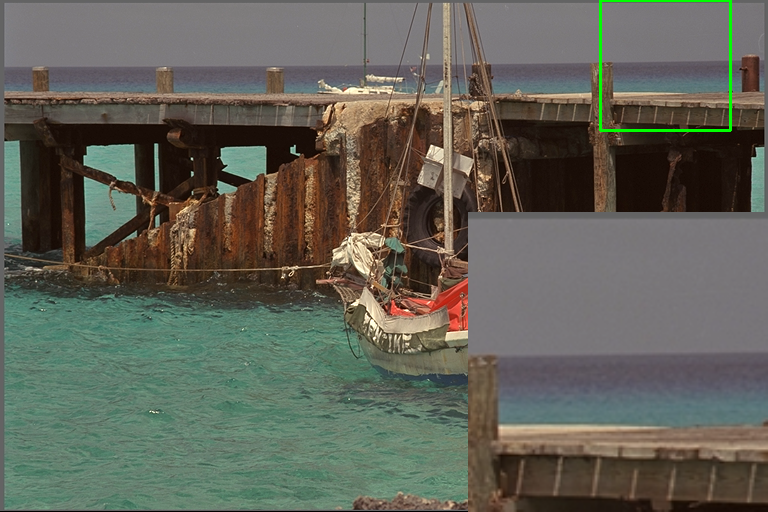}		
		\caption*{\centering{Original Image~\linebreak{(SSIM/W-dis)}}}
		\label{fig:10hbd}
	\end{subfigure}
	\begin{subfigure}{0.32\linewidth}
		\centering		
		\includegraphics[width=\linewidth]{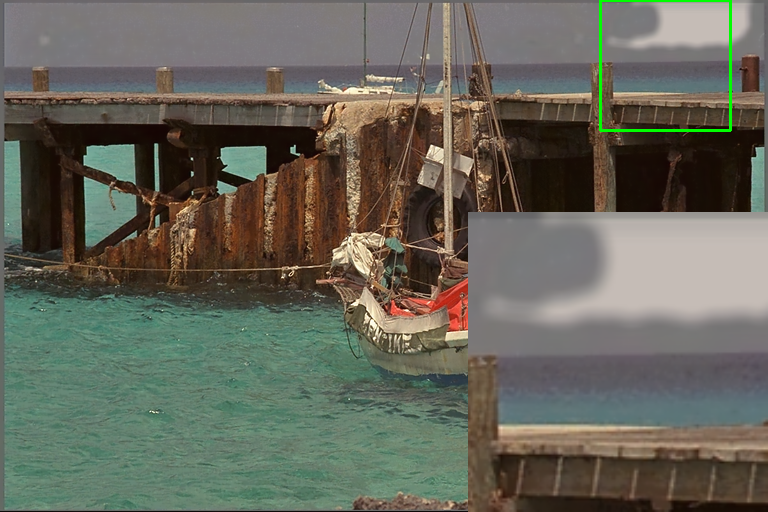}		
		\caption*{\centering{L-BDEN~\cite{lbden_tcsvt2021}~\linebreak{(0.9698/1.10)}}}
		\label{fig:10lbden}
	\end{subfigure}
	\begin{subfigure}{0.32\linewidth}
		\centering		
		\includegraphics[width=\linewidth]{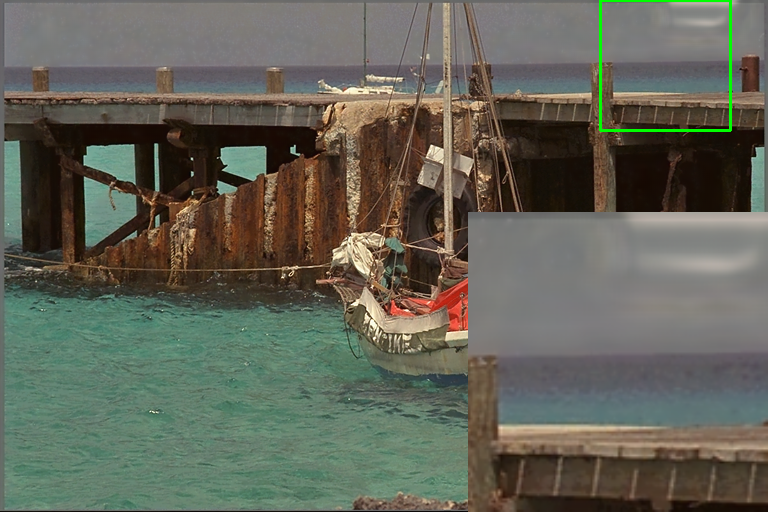}		
		\caption*{\centering{RMFNet~\cite{rmfnet_csvt2021}~\linebreak{(0.9679/0.67)}}}
		\label{fig:10rmf}
	\end{subfigure}

	\caption{Two examples of PSNR/SSIM and W-Dis comparisons from restoring 4-bit to 8-bit.}
	\label{fig:wdis_example}
\end{figure}

To further investigate the superiority of W-dis than traditional metrics, we choose two representative examples to compare PSNR/SSIM and W-dis, as shown in \Cref{fig:wdis_example}. In the first row of the figure, we focus on the color of the ship. The result of BE-CALF is more different to the original image than the result of RMFNet. Correspondingly, although the PSNR of BE-CALF is higher than RMFNet, the W-dis of RMFNet is lower. Similar situation can be observed in the second row of \Cref{fig:wdis_example}. In the sky area, we can find that L-BDEN generates more artifacts than the RMFNet. Although the SSIM of L-BDEN is higher, the W-dis of RMFNet is lower, and the perceptual experience is better. In this point of view, W-dis can well describe the perceptual difference between two images.

\section{Experiment}
We train and test our BRNet on five widely used datasets: DIV2K~\cite{div2k_cvprw2017}, Kodak~\cite{kodim}, Set5~\cite{set5}, Set14~\cite{set14}, and B100~\cite{b100}. We train our method on 900 images from DIV2K for SDR restoration. DIV2K is a high-resolution color image dataset, and has been widely used for image restoration tasks. For test, we choose 5 images from Set5, 14 images from Set14, 100 images from B100, and 24 images from Kodak datasets seperately for comparing the performances of models. We use Adam optimizer~\cite{adam_iclr2015} to update the model with L1-loss, and halve the learning rate for every $200$ epochs. We choose PSNR, SSIM and Wasserstein distance (W-dis) as metrics.

\begin{table}[t]
	\scriptsize      
	\centering
	
	\begin{tabular}{|c|c|c|c|c|c|c|}
		\hline
		\multirow{2}*{\textbf{BD}}& \multicolumn{2}{c|}{\textbf{PSNR$\uparrow$}}& \multicolumn{2}{c|}{\textbf{SSIM$\uparrow$}}& \multicolumn{2}{c|}{\textbf{W-dis$\downarrow$}}\\
		\cline{2-7}
		& w PT& w/o PT& w PT& w/o PT& w PT& w/o PT\\
		\hline
		\hline
		\textbf{1}& 19.14 & \textbf{19.15} & \textbf{0.6854} & 0.6827 & \textbf{13.68} & 14.02 \\
		\textbf{2}& \textbf{27.91} & 27.71 & 0.8779 & \textbf{0.8784} & \textbf{3.03} & 3.63 \\
		\textbf{3}& \textbf{34.22} & 34.03 & \textbf{0.9495} & 0.9489 & \textbf{1.22} & 1.41 \\
		\textbf{4}& \textbf{39.80} & 39.61 & \textbf{0.9794} & 0.9789 & \textbf{0.59} & 0.64 \\
		\hline
	\end{tabular}
	
	\caption{Performance with or without progressive training on different bit-depth recovery on Kodak dataset. \textbf{BD} denotes the bit depth to be replenished. The best performance is labeled in \textbf{bold}.}
	\label{tab:abl-cl}
\end{table}

\subsection{Model Analysis}

\subsubsection{Analysis on Progressive Training Strategy}
To validate the performance of progressive training (PT), we train the network with and without PT and compare the performances, respectively. The training data without PT are randomly dropped different number of bits from $1$ to $7$, and all other settings of the two models are totally the same. \cref{fig:cl} shows the training loss comparison between the two training strategies. The red curve is the loss variation without PT and the blue one is the loss changes with PT. The dash lines in the figure note the converged losses of the two models. In the figure, we can find the loss with PT rises in the first 60 epochs. This is because we increase the difficulty according to the progressive training. The converged loss is calculated by averaging the losses of the last 20 epochs. The converged L1-loss of model with PT is 5.22 which is lower than 5.30 without PT. In this point of view, the blue dash line is lower than the red one, indicating the proposed PT improves the capacity.

\cref{tab:abl-cl} shows the quantitative comparisons with and without PT in terms of PSNR, SSIM and {W-dis} on Kodak dataset. In the table, we compare the results of restoring 1, 2, 3, and 4 bits to 8-bit. In all situations, the model trained with PT achieves better W-dis than the model without PT. When the expended bit depth is 1, the model with PT achieves competitive performance to the model without PT. In other situations, the model with PT surpasses the model without PT near 0.2 dB in average on PSNR, separately. 

\begin{table}[t]
	\scriptsize      
	\centering
	
	\begin{tabular}{|c|c|c|c|c|c|c|}
		\hline
		\multirow{2}*{\textbf{BD}}& \multicolumn{2}{c|}{\textbf{PSNR$\uparrow$}}& \multicolumn{2}{c|}{\textbf{SSIM$\uparrow$}}& \multicolumn{2}{c|}{\textbf{W-dis$\downarrow$}}\\
		\cline{2-7}
		&Weight& Value & Weight& Value& Weight& Value\\
		\hline
		\hline
		\textbf{1}& \textbf{19.14} & 18.97 & \textbf{0.6854} & 0.6766 & \textbf{13.68} & 14.09 \\
		\textbf{3}& \textbf{34.22} & 33.55 & \textbf{0.9495} & 0.9469 & \textbf{1.22} & 1.65 \\
		\textbf{5}& \textbf{44.32} & 43.43 & \textbf{0.9909} & 0.9904 & \textbf{0.32} & 0.54 \\
		\textbf{7}& \textbf{52.90} & 47.07 & \textbf{0.9998} & 0.9978 & \textbf{0.06} & 0.99 \\
		\hline
	\end{tabular}
	
	\caption{Performance comparison between learning weighting map (labeled as ``Weight'') and directly learning values (labeled as ``Value'') on Kodak dataset. \textbf{BD} denotes the bit depth to be replenished. The best performance is labeled in \textbf{bold}.}
	\label{tab:abl-contour}
\end{table}

\begin{figure}[t]
	\centering
	\begin{subfigure}{0.3\linewidth}
		\centering		
		\includegraphics[width=\linewidth]{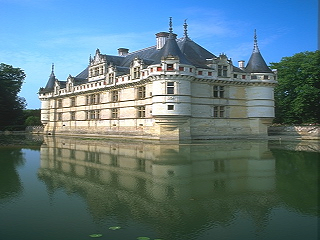}		
		\caption{Original Image}
		\label{fig:hbd}
	\end{subfigure}
	\begin{subfigure}{0.3\linewidth}
		\centering				
		\includegraphics[width=\linewidth]{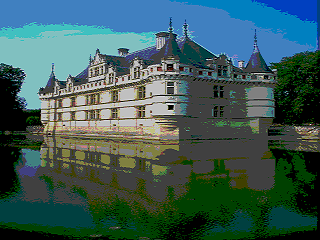}		
		\caption{LBD input}
		\label{fig:lbd}
	\end{subfigure}
	\begin{subfigure}{0.3\linewidth}
		\centering		
		\includegraphics[width=\linewidth]{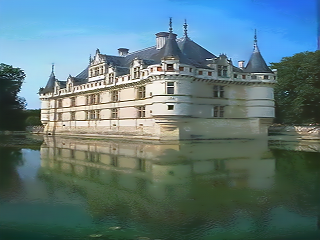}		
		\caption{Restored image}
		\label{fig:sbd}
	\end{subfigure}
	\begin{subfigure}{0.3\linewidth}
		\centering		
		\includegraphics[width=\linewidth]{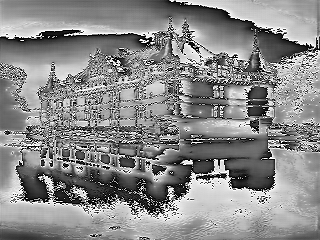}		
		\caption{Red Channel}
		\label{fig:colorR}
	\end{subfigure}
	\begin{subfigure}{0.3\linewidth}
		\centering		
		\includegraphics[width=\linewidth]{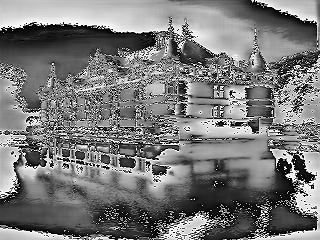}		
		\caption{Green Channel}
		\label{fig:colorG}
	\end{subfigure}
	\begin{subfigure}{0.3\linewidth}
		\centering		
		\includegraphics[width=\linewidth]{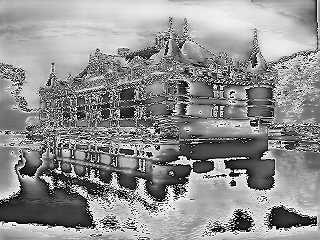}		
		\caption{Blue Channel}
		\label{fig:colorB}
	\end{subfigure}
	\caption{An example of learned weighting map on restoring 2-bit image to 8-bit. (a): Original image. (b) The 2-bit LBD input. (c) The restored 8-bit image. (d)-(f): weighting maps of RGB channels separately. The darker color means lower weight value.
	}
	\label{fig:vis_res}
\end{figure}

\subsubsection{Analysis on Weighting Map Learning}
To validate the effectiveness of learning weighting map for replenishment, we modify BRNet by removing the last Sigmoid layer and the input $\N_b$, which makes the network the same as other CNNs and directly learns desired values. We keep other settings unchanged so that the modified network is trained in all bit-depth situations. \cref{tab:abl-contour} shows the PSNR/SSIM/W-dis comparison on Kodak dataset between learning weight and learning direct values. ``Weight'' indicates the weighting learning manner in the paper, while ``Value'' denotes the modified version. In the table, the weighting map learning outperforms the modified version in all situations. The PSNR of ``Weight'' is 44.32 dB when bit depth is 5, which is near 1 dB higher than that of the ``Value'' manner. When the bit-depth is 7, ``Weight'' achieves near 6 dB PSNR improvement than ``Value''. 

We also visualize the learned weighting map to investigate the effectiveness of our network. \cref{fig:vis_res} shows the learned weighting maps of RGB channels for replenishment. The darker color means the lower weight value. In \cref{fig:sbd}, the blue channel plays an important role with higher value in the sky region, while the red channel and green channel reconcile the pure blue color to a gradual changes in the restored image. As such, the learned weighting maps can adaptively find the missing information from LBD images and generate a natural color transaction image.

\begin{table*}[t]
	\centering
	
	\begin{tabular}{|c|c|cc|cccc|c|}
		\hline
		\textbf{Bit-Depth}& \textbf{Indicator}& \textbf{Zero Padding}& \textbf{IPAD~\cite{ipad_tip2018}}& \textbf{BitNet~\cite{bitnet_accv2018}}&  \textbf{LBDEN~\cite{lbden_tcsvt2021}}& \textbf{BE-CALF~\cite{be_calf_tip2019}}& \textbf{RMFNet~\cite{rmfnet_csvt2021}} 
		&  \textbf{BRNet} \\
		
		\hline \hline
		\multirow{3}*{\textbf{1}}
		& \textbf{PSNR$\uparrow$}   & 10.79&  17.51&  12.03&  17.61&  18.11&  18.31     
		&  \textbf{19.14} \\
		
		& \textbf{SSIM$\uparrow$}   & 0.3067& 0.5451& 0.3946& 0.6090& 0.6383& 0.6370    
		& \textbf{0.6854} \\
		
		& \textbf{W-dis$\downarrow$}& 64.82&  17.84&  53.77&  18.45&  17.06&  16.99     
		&   \textbf{13.68}\\
		
		\hline
		\multirow{3}*{\textbf{3}}
		& \textbf{PSNR$\uparrow$}   & 22.77&  29.20&  32.68&  32.09&  32.56&  32.05     
		&  \textbf{34.22} \\
		
		& \textbf{SSIM$\uparrow$}   & 0.8559& 0.9000& 0.9339& 0.9355& 0.9360& 0.9294    
		& \textbf{0.9495} \\
		
		& \textbf{W-dis$\downarrow$}& 16.00&   2.59&  2.2883&  1.96&   2.24&   1.79     
		&   \textbf{1.22}\\
		
		\hline
		\multirow{3}*{\textbf{4}}
		& \textbf{PSNR$\uparrow$}   & 29.06&  34.90&  38.48&  38.20&  38.44&  37.75     
		&  \textbf{39.80} \\
		
		& \textbf{SSIM$\uparrow$}   & 0.9484& 0.9595& 0.9740& 0.9750& 0.9743& 0.9712    
		& \textbf{0.9794} \\
		
		& \textbf{W-dis$\downarrow$}& 7.68&   1.11&   1.07&   0.83&   1.02&   0.79      
		&      \textbf{0.59}\\
		
		\hline
		\multirow{3}*{\textbf{5}}
		& \textbf{PSNR$\uparrow$}   & 35.55&  40.35&  43.30&  43.08&  43.51&  42.63     
		&  \textbf{44.32} \\
		
		& \textbf{SSIM$\uparrow$}   & 0.9839& 0.9843& 0.9887& 0.9896& 0.9898& 0.9882    
		& \textbf{0.9909} \\
		
		& \textbf{W-dis$\downarrow$}& 3.50&   0.69&   0.60&   0.45&   0.59&   0.45      
		&   \textbf{0.32}\\
		
		\hline
		\multirow{3}*{\textbf{7}}
		& \textbf{PSNR$\uparrow$}   & 51.02&  51.02&  47.13&  45.74&  48.79&  47.66     
		&  \textbf{52.90} \\
		
		& \textbf{SSIM$\uparrow$}   & 0.9985& \textbf{0.9985}& 0.9978& 0.9971& 0.9976& 0.9974       
		& {0.9982} \\
		
		& \textbf{W-dis$\downarrow$}& 0.52&      0.51&   0.86&   0.62&   0.61&   0.36   
		&   \textbf{0.03}\\
		
		\hline		
	\end{tabular}
	
	\caption{Performance comparison with state-of-the-art methods on restoring image with different bit depth on Kodak dataset. The \textbf{bold} value means the best performance.}
	\label{tab:bit1357}
\end{table*}

\begin{figure*}[t]
	\centering
	
	\begin{subfigure}{0.18\linewidth}
		\centering		
		\includegraphics[width=\linewidth]{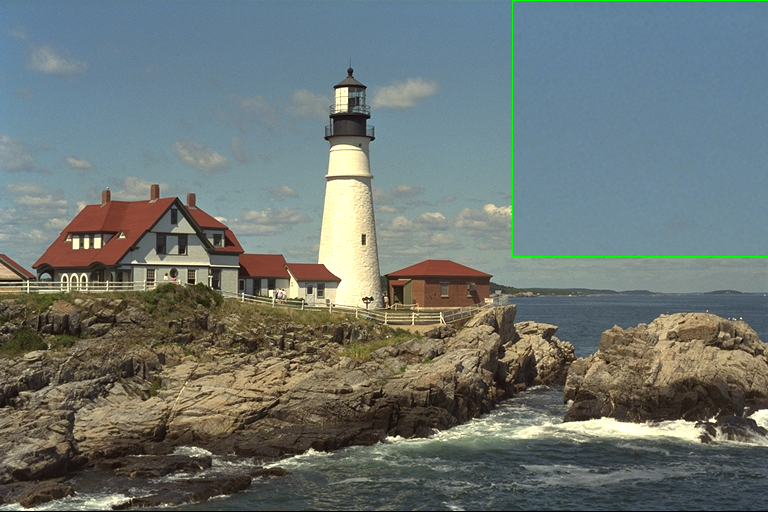}		
		\caption*{\centering{Original Image~\linebreak{(PSNR/SSIM/W-dis)}}}
	\end{subfigure}
	\begin{subfigure}{0.18\linewidth}
		\centering		
		\includegraphics[width=\linewidth]{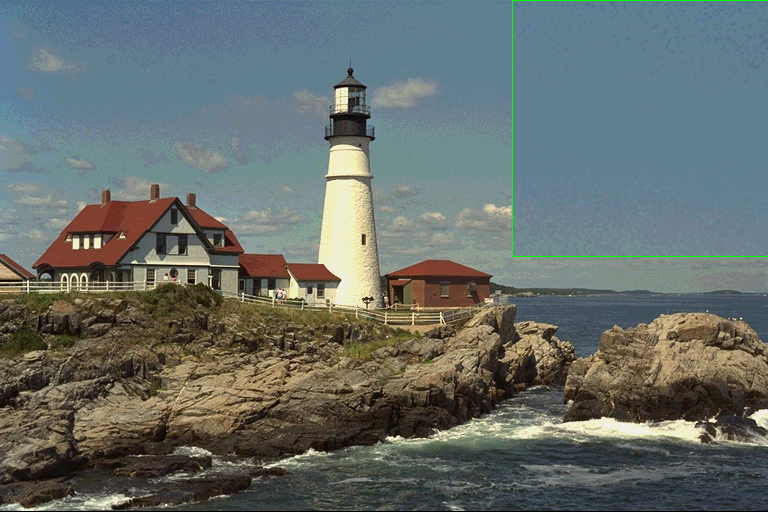}		
		\caption*{\centering{Zero Padding~\linebreak{(29.21 / 0.9491 / 7.53)}}}
	\end{subfigure}
	\begin{subfigure}{0.18\linewidth}
		\centering		
		\includegraphics[width=\linewidth]{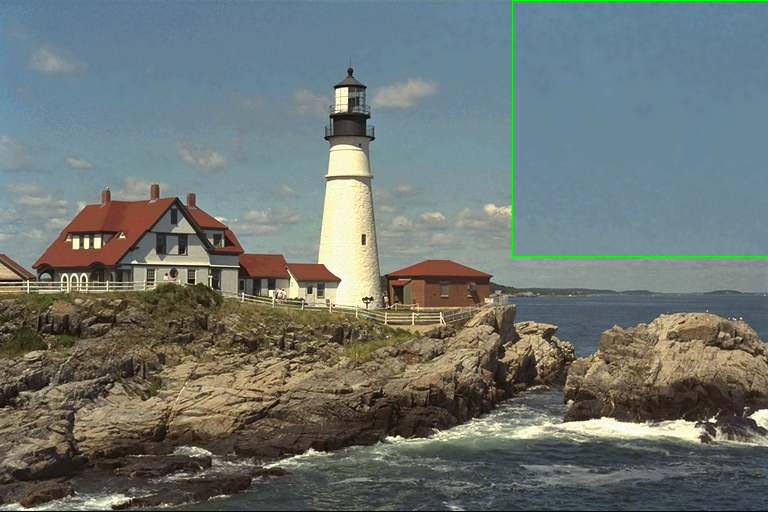}		
		\caption*{\centering{BE-CALF~\cite{be_calf_tip2019}~\linebreak{(39.02 / 0.9732 / 0.81)}}}
	\end{subfigure}
	\begin{subfigure}{0.18\linewidth}
		\centering		
		\includegraphics[width=\linewidth]{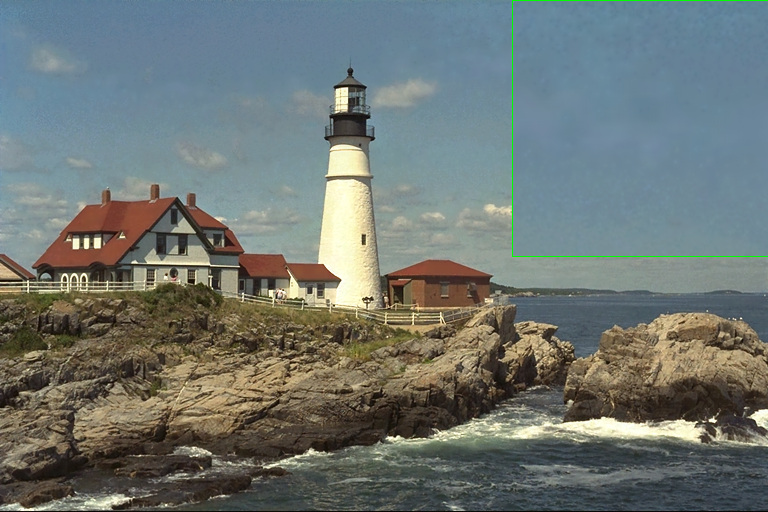}		
		\caption*{\centering{RMFNet~\cite{rmfnet_csvt2021}~\linebreak{(38.22 / 0.9708 / 0.66)}}}
	\end{subfigure}
	\begin{subfigure}{0.18\linewidth}
		\centering		
		\includegraphics[width=\linewidth]{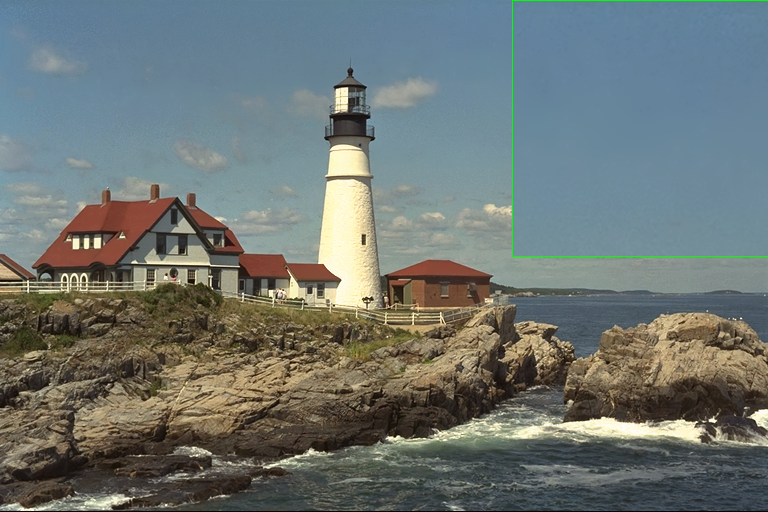}		
		\caption*{\centering{BRNet~\linebreak{(\textbf{40.12 / 0.9778 / 0.43})}}}
	\end{subfigure}	
	
	\caption{Visual comparisons on restoring from 4-bit to 8-bit. Please magnify the result for a better view.}
	\label{fig:vis}
\end{figure*}

\begin{figure*}[t]
	\centering
	\begin{subfigure}{0.18\linewidth}
		\centering		
		\includegraphics[width=\linewidth]{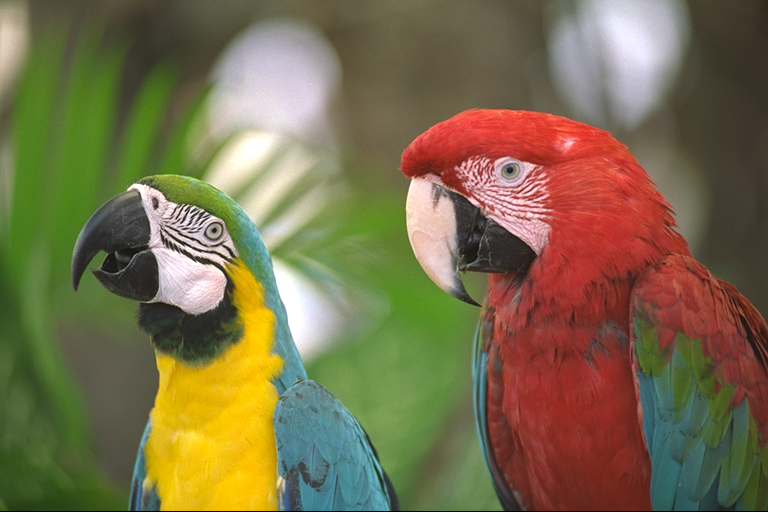}		
		\caption*{\centering{Original Image~\linebreak{(PSNR/SSIM/W-dis)}}}
	\end{subfigure}
	\begin{subfigure}{0.18\linewidth}
		\centering		
		\includegraphics[width=\linewidth]{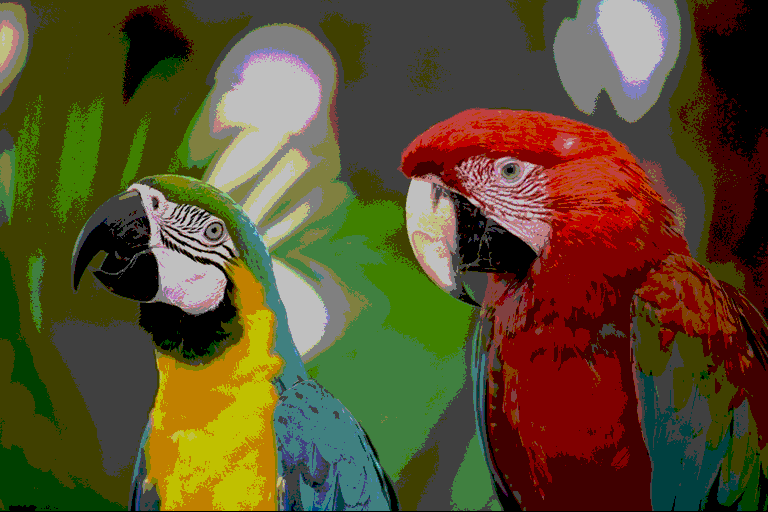}		
		\caption*{\centering{Zero Padding~\linebreak{(16.67 / 0.6604 / 32.41)}}}
	\end{subfigure}
	\begin{subfigure}{0.18\linewidth}
		\centering		
		\includegraphics[width=\linewidth]{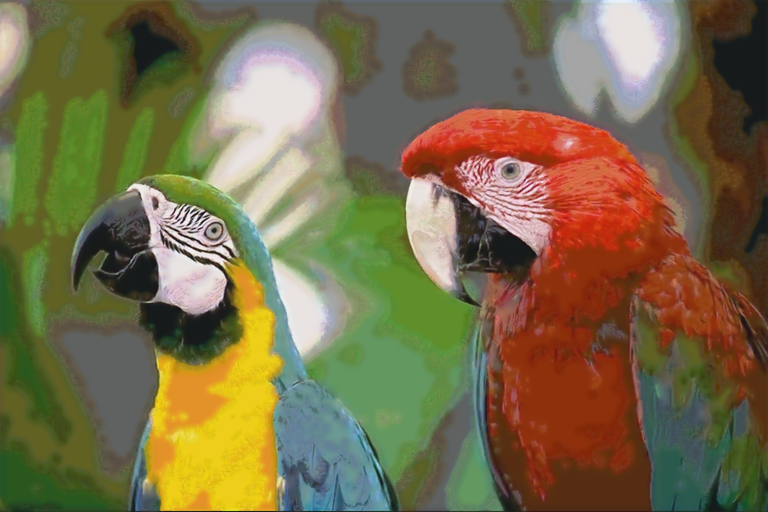}		
		\caption*{\centering{BE-CALF~\cite{be_calf_tip2019}~\linebreak{(25.42 / 0.8515 / 6.24)}}}
	\end{subfigure}
	\begin{subfigure}{0.18\linewidth}
		\centering		
		\includegraphics[width=\linewidth]{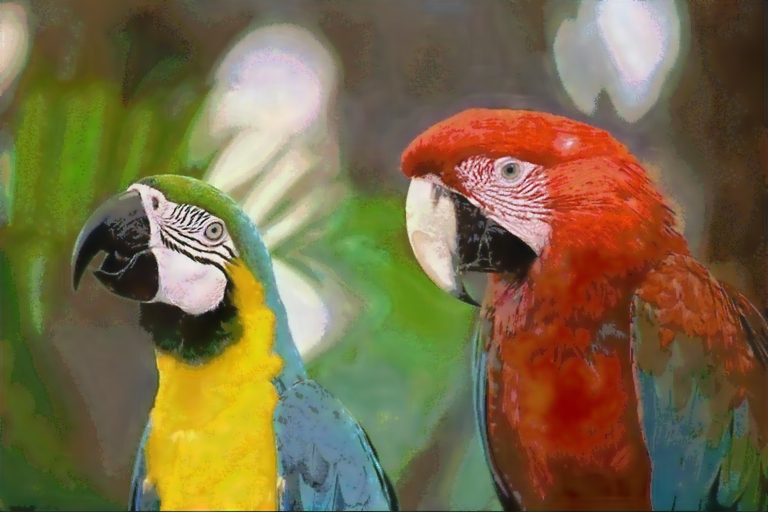}		
		\caption*{\centering{RMFNet~\cite{rmfnet_csvt2021}~\linebreak{(25.33 / 0.8589 / 5.39)}}}
	\end{subfigure}
	\begin{subfigure}{0.18\linewidth}
		\centering		
		\includegraphics[width=\linewidth]{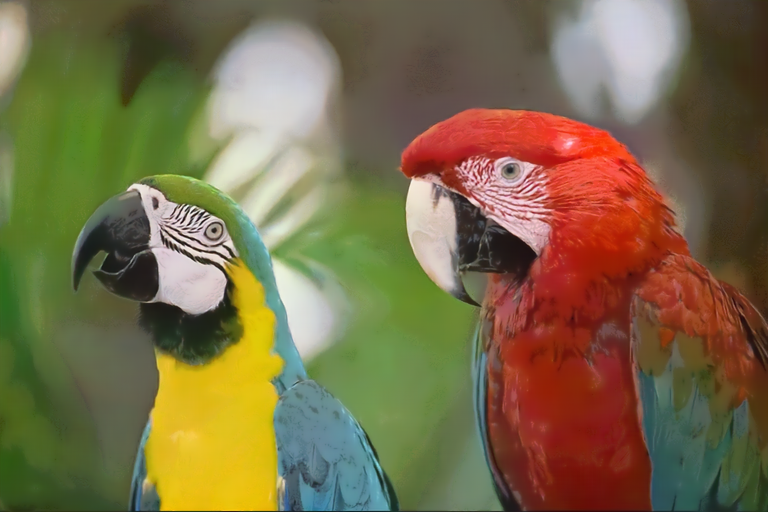}		
		\caption*{\centering{BRNet~\linebreak{(\textbf{27.18 / 0.9037 / 3.71})}}}
	\end{subfigure}
	
	\begin{subfigure}{0.18\linewidth}
		\centering		
		\includegraphics[width=\linewidth]{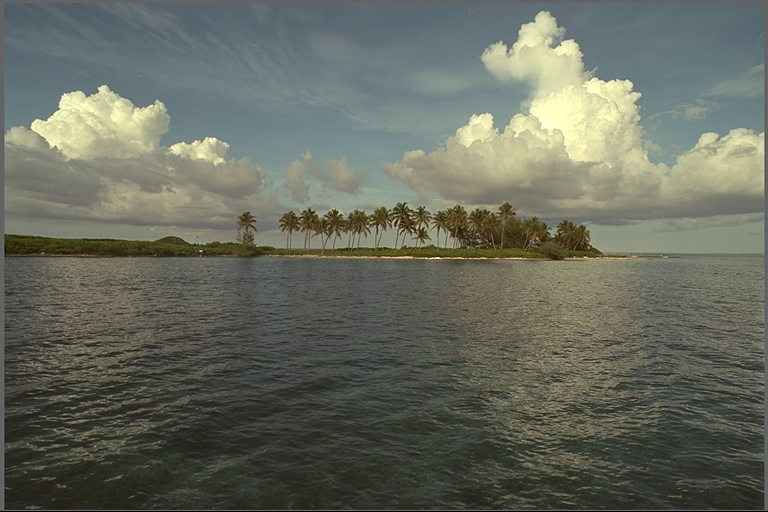}		
		\caption*{\centering{Original Image~\linebreak{(PSNR/SSIM/W-dis)}}}
	\end{subfigure}
	\begin{subfigure}{0.18\linewidth}
		\centering		
		\includegraphics[width=\linewidth]{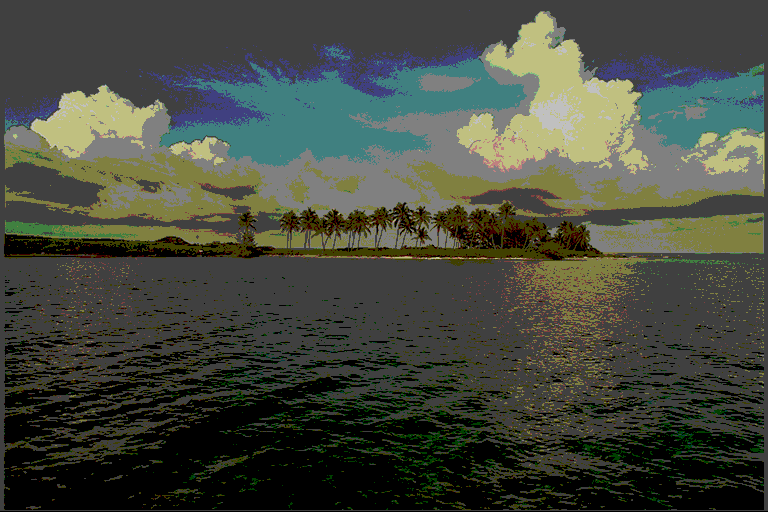}		
		\caption*{\centering{Zero Padding~\linebreak{(16.41 / 0.5969 / 33.74)}}}
	\end{subfigure}
	\begin{subfigure}{0.18\linewidth}
		\centering		
		\includegraphics[width=\linewidth]{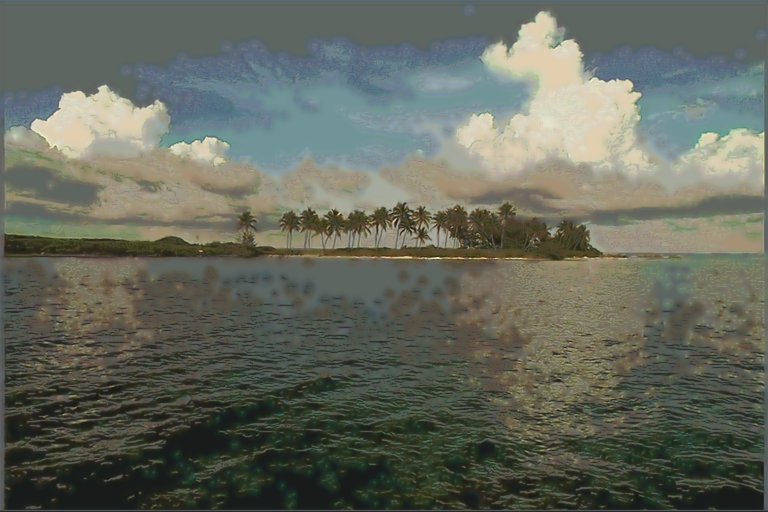}		
		\caption*{\centering{BE-CALF~\cite{be_calf_tip2019}~\linebreak{(26.94 / 0.8175 / 5.28)}}}
	\end{subfigure}
	\begin{subfigure}{0.18\linewidth}
		\centering		
		\includegraphics[width=\linewidth]{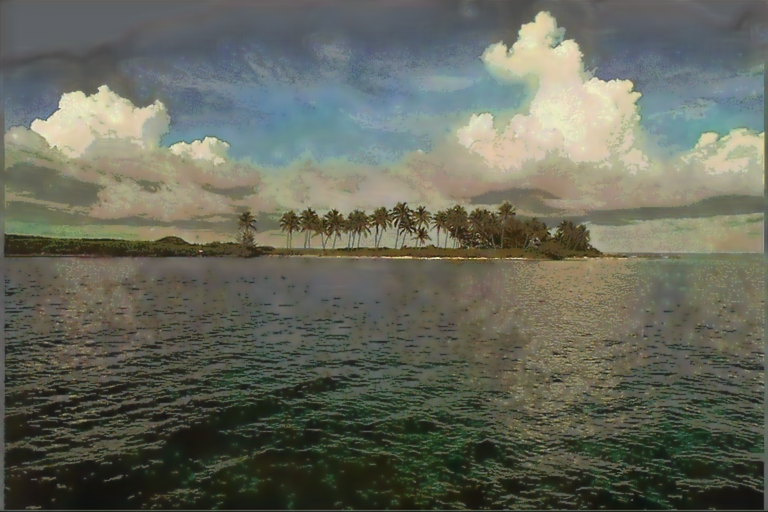}		
		\caption*{\centering{RMFNet~\cite{rmfnet_csvt2021}~\linebreak{(26.03 / 0.8045 / 6.61)}}}
	\end{subfigure}
	\begin{subfigure}{0.18\linewidth}
		\centering		
		\includegraphics[width=\linewidth]{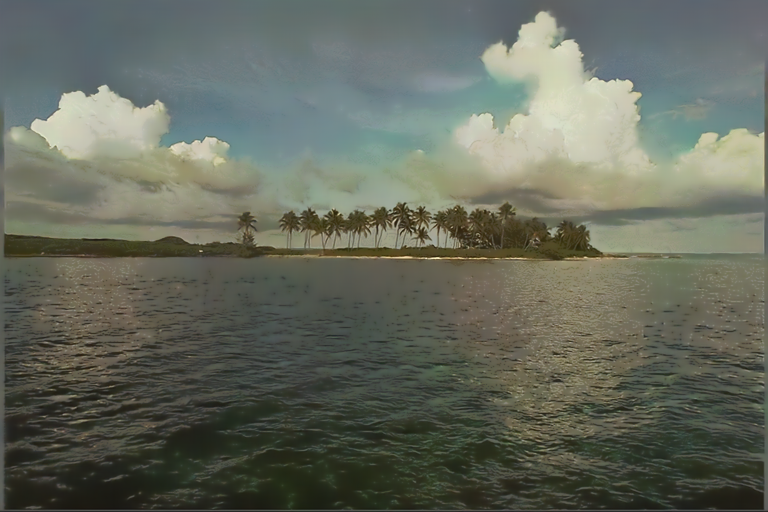}		
		\caption*{\centering{BRNet~\linebreak{(\textbf{28.75 / 0.8511 / 3.26})}}}
	\end{subfigure}	
	
	\caption{Visual comparisons on restoring from 2-bit to 8-bit. Please zoom-up for a better view.}
	\label{fig:wdis}
\end{figure*}

\begin{figure}[t]
	\centering
	\begin{subfigure}{0.32\linewidth}
		\centering		
		\includegraphics[width=\linewidth]{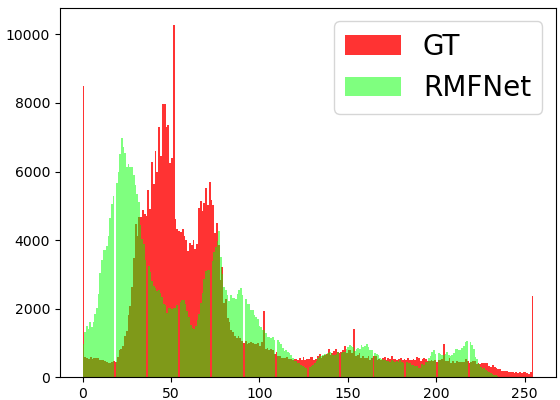}		
	\end{subfigure}
	\begin{subfigure}{0.32\linewidth}
		\centering		
		\includegraphics[width=\linewidth]{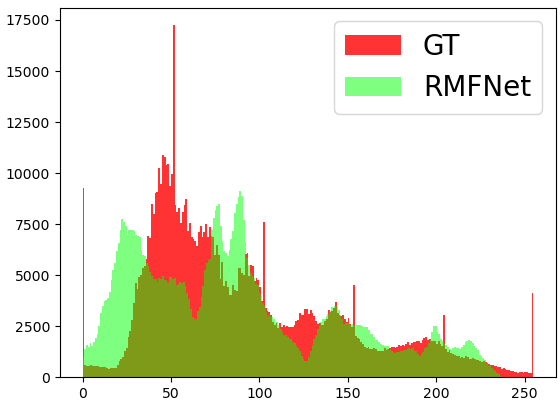}		
	\end{subfigure}
	\begin{subfigure}{0.32\linewidth}
		\centering		
		\includegraphics[width=\linewidth]{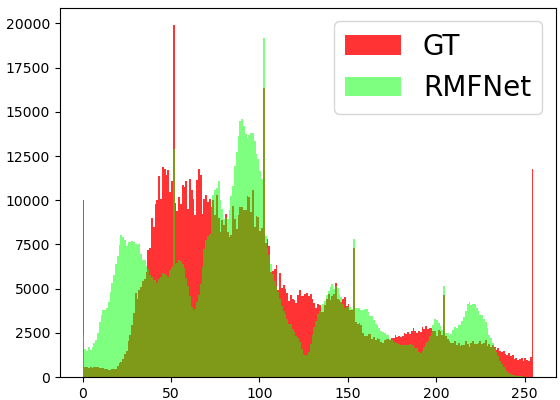}		
	\end{subfigure}
	
	\begin{subfigure}{0.32\linewidth}
		\centering		
		\includegraphics[width=\linewidth]{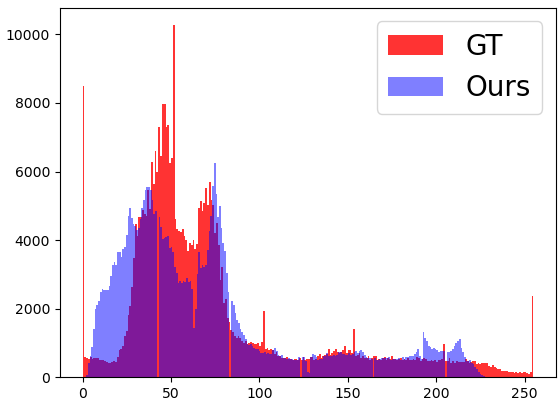}		
		\caption{Blue Channel}
	\end{subfigure}
	\begin{subfigure}{0.32\linewidth}
		\centering		
		\includegraphics[width=\linewidth]{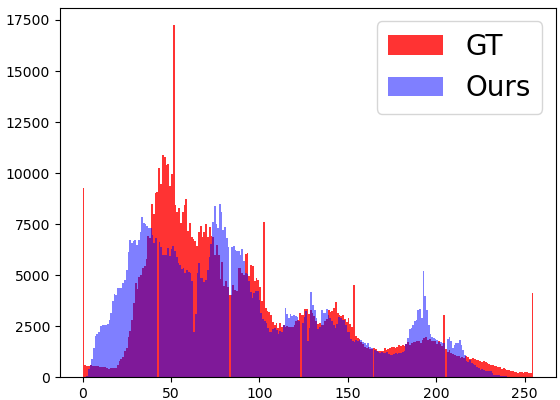}		
		\caption{Green Channel}
	\end{subfigure}
	\begin{subfigure}{0.32\linewidth}
		\centering		
		\includegraphics[width=\linewidth]{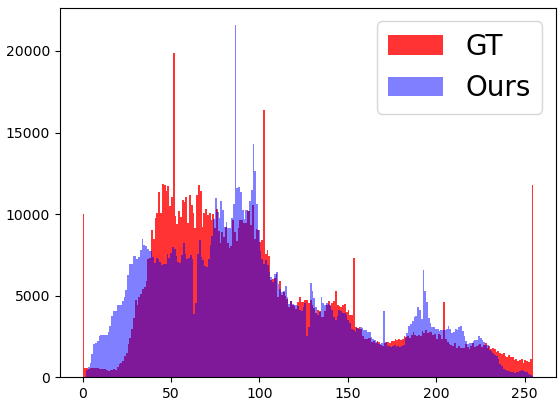}		
		\caption{Red Channel}
	\end{subfigure}
	\caption{Color histogram comparison between RMFNet and ours on restoring 2-bit to 8-bit. Our color distributions are closer to the groundtruth.}
	\label{fig:hist}
\end{figure}


\subsection{Comparison with State-of-the-Art Methods}
We compare our BRNet with 6 state-of-the-art methods, including a recent traditional method (IPAD~\cite{ipad_tip2018}) and five CNN-based methods: BitNet~\cite{bitnet_accv2018}, LBDEN~\cite{lbden_tcsvt2021}, BE-CALF~\cite{be_calf_tip2019}, and RMFNet~\cite{rmfnet_csvt2021}. It should be noticed that vanilla RMFNet, BE-CALF and LBDEN have no solution for different bit depths. For a fair comparison, we re-train these models with our whole training datasets covering the same bit expansion situations with our model.

Following the comparison in previous works, we first investigate the performances of state-of-the-art methods on restoring 4-bit image to 8-bit on Kodak dataset. \cref{tab:bit1357} shows the PSNR/SSIM/W-dis results on Kodak dataset. The best results are shown in bold. ``Zero Padding'' means to set the missing bits as zero. In the table, our method achieves the best performance than other works in all evaluation terms. Specifically, our network gains 0.3 dB higher than BitPlane and 2 dB higher than RMFNet in term of PSNR. The W-dis of our method is 0.2 lower than BitPlane and RMFNet, which indicates a better visual performance. 

\cref{fig:vis} shows the visual comparisons with two competitive methods on restoring 4-bit image to 8-bit. As can be observed, BRNet generates more natural textures on the sky and cloud while other works suffer from artifacts and false contours. In the zoomed-in sky area, our result contains fewer noises and produces more correct colors than other works, which is the closest to the ground truth. It is worth noting that the PSNR/SSIM results of RMFNet is lower than the BE-CALF. In contrast, RMFNet holds better W-dis result than BE-CALF with higher visual quality, indicating W-dis is a suitable metric for visual quality assessment.

To show the general applicability of BRNet, we compare the performances in different bit-depth situations. We set the bit depth as 1, 3, 5, and, 7, and demonstrate the PSNR/SSIM/W-dis results in \cref{tab:bit1357}. In the table, BRNet outperforms the state-of-the-art methods in all bit expansion situations. In general, with the increase of bit depth, the superiority of BRNet becomes more obvious. It should be noticed that BitPlane has no capability on restoring 1-bit image. IPAD is a traditional method that intrinsically satisfies different bit-depth situations. Compared with IPAD, BRNet achieves near 4 dB improvements in term of PSNR when the bit-depth is 5. 

\cref{fig:wdis} shows the visual comparisons on restoring 2-bit to 8-bit. In the figure, our method achieves the best visual performance than other works with superior PSNR/SSIM/W-Dis results. We can find that BRNet can still generate satisfactory results when suffering severe information loss. In the flat area where banding effect is obvious, such as the background of the parrot and the sky over sea, BRNet can remove the false contours and estimate the correct colors and textures more effectively.

To further investigate the effectiveness of BRNet, we compare the color distributions restored by our network and RMFNet respectively, as shown in \cref{fig:hist}. The histograms are constructed in red, blue and green channels respectively. The distributions recovered by our network are much close to the ground truth, while the histograms of RMFNet has huge difference with ground truth. The overlapping area between our distribution and the ground truth is larger than that of RMFnet, which explains why we achieve the best W-dis performance.

We also compare the performances on Set5, Set14, and BSD100 datasets, where BRNet outperforms other works near 2 dB, 3 dB, 1.5 dB separately in term of PSNR on restoring 4-bit image to 8-bit. Table \ref{tab:set5}, \ref{tab:set14} and \ref{tab:b100} shows the PSNR~\cite{psnr}/SSIM~\cite{ssim}/W-dis~\cite{wasserstein_aistas2021} comparisons on different datasets separately. We can find that our BRNet achieves the best performance on all testing benchmarks with different settings of bit depth. With the increase of bit depth, BRNet outperforms other works with much higher PSNR results. Specially, when the bit-depth is 7, our BRNet gains more than 5 dB, 10 dB, and 3 dB PSNR improvement than other works on Set5, Set14, and B100 datasets separately. 

\begin{figure*}[t]
	\centering
	
	\begin{subfigure}{0.16\linewidth}
		\centering		
		\includegraphics[width=\linewidth]{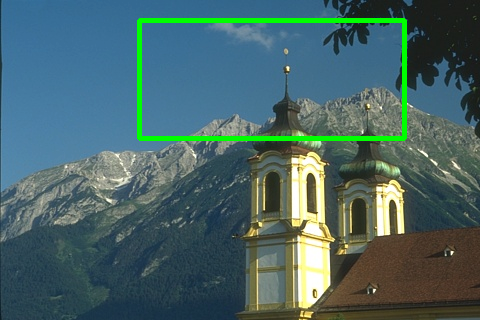}		
		\caption*{\centering{Original Image~\linebreak{(PSNR/SSIM/W-dis)}}}
	\end{subfigure}
	\begin{subfigure}{0.16\linewidth}
		\centering		
		\includegraphics[width=\linewidth]{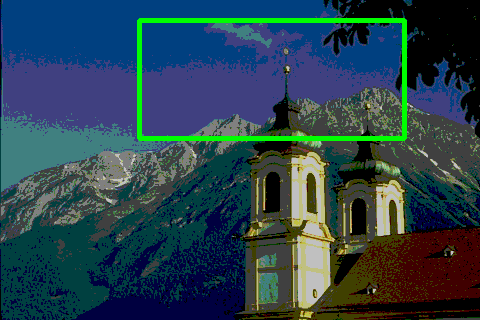}		
		\caption*{\centering{Zero Padding~\linebreak{(16.78/0.6400/31.48)}}}
	\end{subfigure}
	\begin{subfigure}{0.16\linewidth}
		\centering		
		\includegraphics[width=\linewidth]{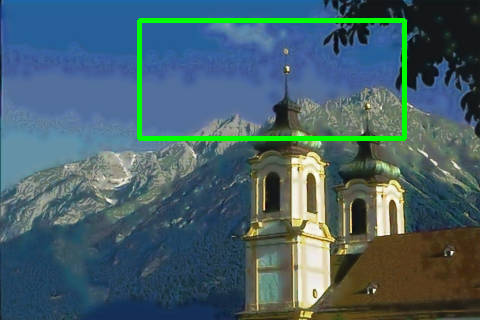}		
		\caption*{\centering{BE-CALF~\cite{be_calf_tip2019}~\linebreak{(26.84/0.8816/6.13)}}}
	\end{subfigure}
	\begin{subfigure}{0.16\linewidth}
		\centering		
		\includegraphics[width=\linewidth]{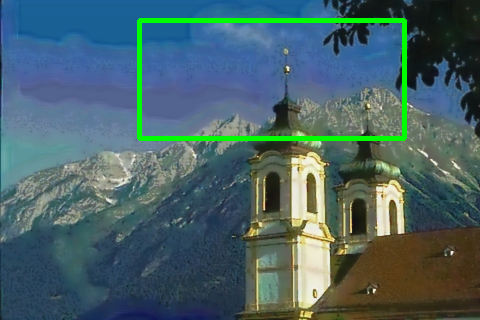}		
		\caption*{\centering{LBDEN~\cite{lbden_tcsvt2021}~\linebreak{(26.41/0.8807/5.62)}}}
	\end{subfigure}
	\begin{subfigure}{0.16\linewidth}
		\centering		
		\includegraphics[width=\linewidth]{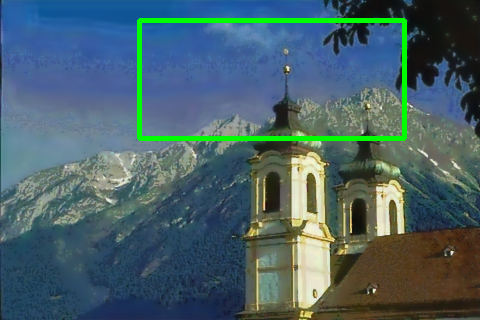}		
		\caption*{\centering{RMFNet~\cite{rmfnet_csvt2021}~\linebreak{(25.74/0.8756/6.28)}}}
	\end{subfigure}
	\begin{subfigure}{0.16\linewidth}
		\centering		
		\includegraphics[width=\linewidth]{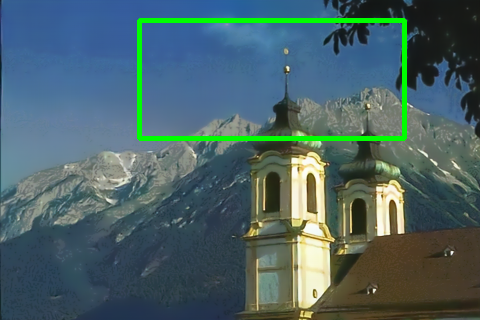}		
		\caption*{\centering{BRNet~\linebreak{(\textbf{28.74/0.9063/2.29})}}}
	\end{subfigure}	
	
	\begin{subfigure}{0.16\linewidth}
		\centering		
		\includegraphics[width=\linewidth]{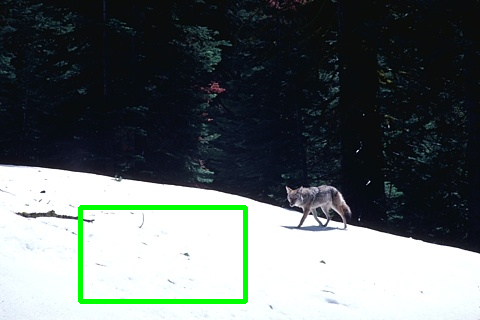}		
		\caption*{\centering{Original Image~\linebreak{(PSNR/SSIM/W-dis)}}}
	\end{subfigure}
	\begin{subfigure}{0.16\linewidth}
		\centering		
		\includegraphics[width=\linewidth]{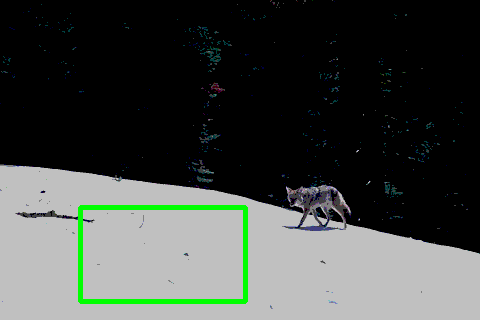}		
		\caption*{\centering{Zero Padding~\linebreak{(16.03/0.3941/32.94)}}}
	\end{subfigure}
	\begin{subfigure}{0.16\linewidth}
		\centering		
		\includegraphics[width=\linewidth]{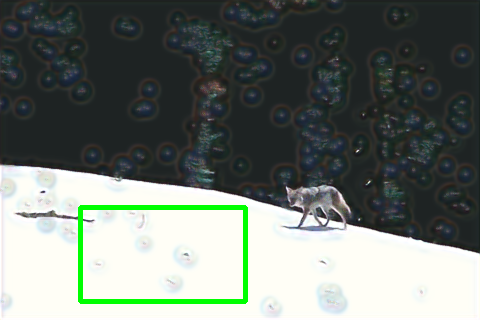}		
		\caption*{\centering{BE-CALF~\cite{be_calf_tip2019}~\linebreak{(21.99/0.6446/15.39)}}}
	\end{subfigure}
	\begin{subfigure}{0.16\linewidth}
		\centering		
		\includegraphics[width=\linewidth]{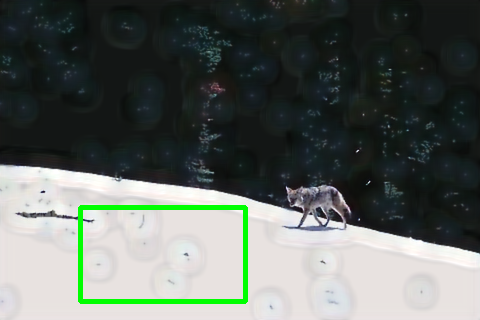}		
		\caption*{\centering{LBDEN~\cite{lbden_tcsvt2021}~\linebreak{(22.59/0.7625/14.37)}}}
	\end{subfigure}
	\begin{subfigure}{0.16\linewidth}
		\centering		
		\includegraphics[width=\linewidth]{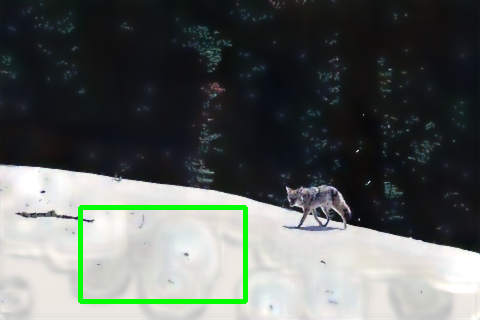}		
		\caption*{\centering{RMFNet~\cite{rmfnet_csvt2021}~\linebreak{(22.34/0.7997/13.10)}}}
	\end{subfigure}
	\begin{subfigure}{0.16\linewidth}
		\centering		
		\includegraphics[width=\linewidth]{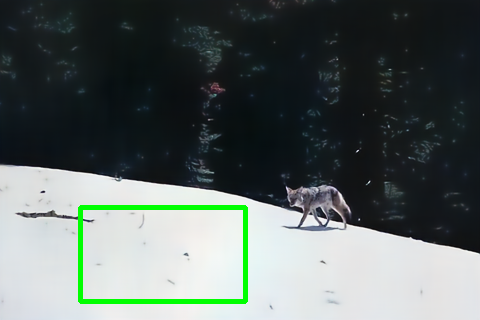}		
		\caption*{\centering{BRNet~\linebreak{(\textbf{23.89/0.8192/11.90})}}}
	\end{subfigure}
	
	\caption{Visual comparisons on restoring images from 2-bit to 8-bit on B100 dataset. Please magnify the result for a better view.}
	\label{fig:vis-b100}
\end{figure*}

Figure \ref{fig:vis-b100} shows the visual comparisons on restoring images from 2-bit to 8-bit on B100 dataset. In the first row of the figure, we can find that our BRNet produces the most satisfactory visual results. Especially in the sky area, BRNet contains fewer false contours and more correct colors than other methods. In the second row, other works are suffering from the false blurs on the snow, while BRNet can effectively suppress the unpleasant artifacts.

\begin{figure*}[t]
	\centering
	
	\begin{subfigure}{0.16\linewidth}
		\centering		
		\includegraphics[width=\linewidth]{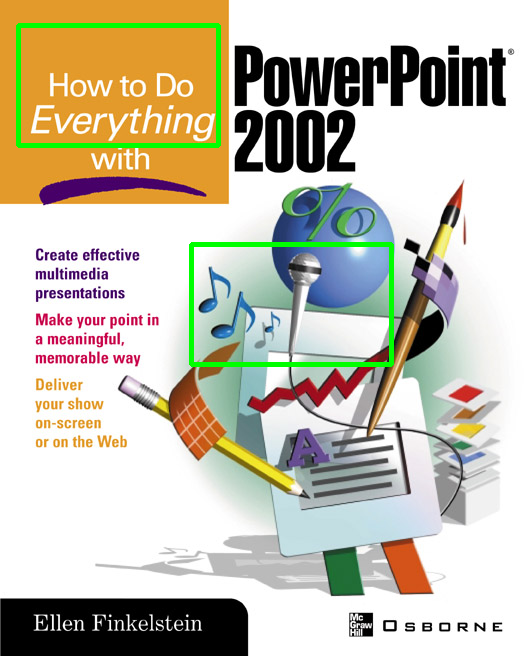}		
		\caption*{\centering{Original Image~\linebreak{(PSNR/SSIM/W-dis)}}}
	\end{subfigure}
	\begin{subfigure}{0.16\linewidth}
		\centering		
		\includegraphics[width=\linewidth]{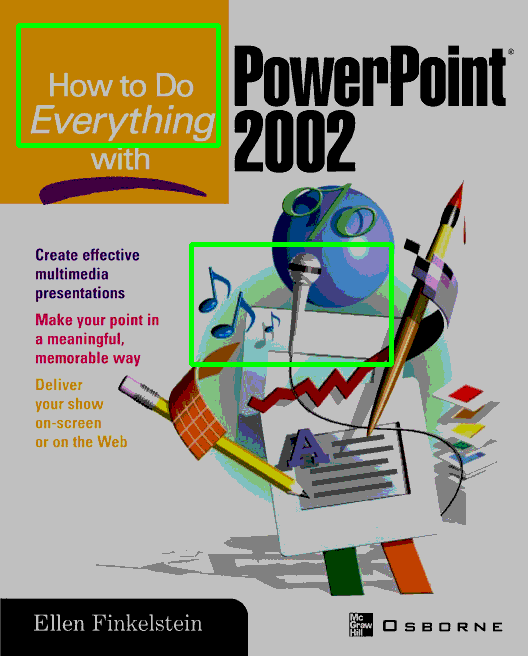}		
		\caption*{\centering{Zero Padding~\linebreak{(14.29/0.9061/44.11)}}}
	\end{subfigure}
	\begin{subfigure}{0.16\linewidth}
		\centering		
		\includegraphics[width=\linewidth]{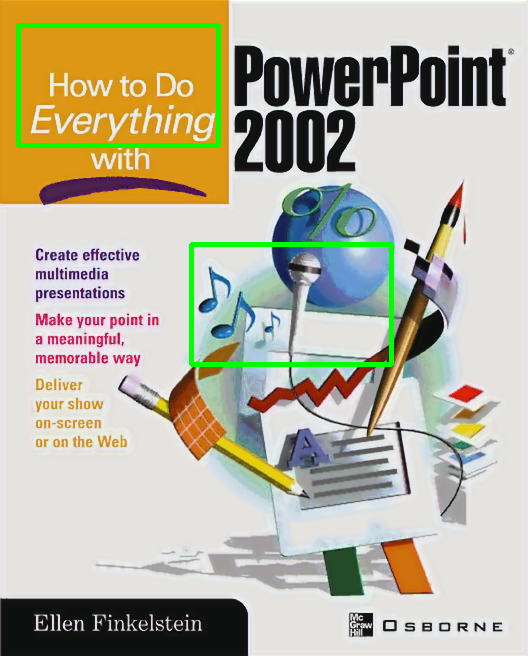}		
		\caption*{\centering{BE-CALF~\cite{be_calf_tip2019}~\linebreak{(21.50/0.9030/16.64)}}}
	\end{subfigure}
	\begin{subfigure}{0.16\linewidth}
		\centering		
		\includegraphics[width=\linewidth]{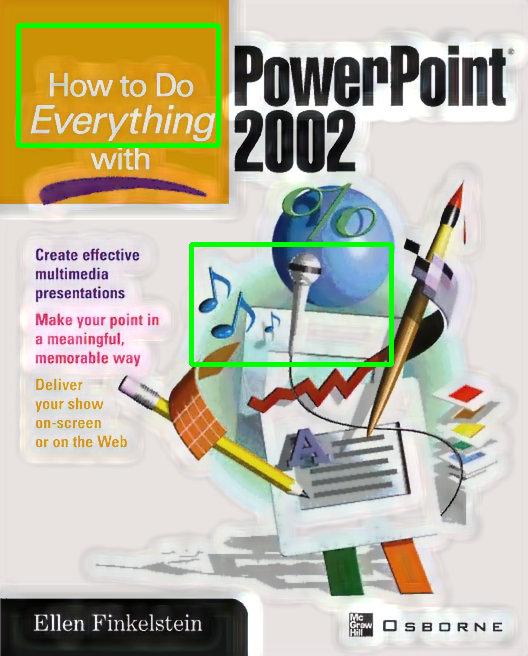}		
		\caption*{\centering{LBDEN~\cite{lbden_tcsvt2021}~\linebreak{(20.52/0.8521/15.64)}}}
	\end{subfigure}
	\begin{subfigure}{0.16\linewidth}
		\centering		
		\includegraphics[width=\linewidth]{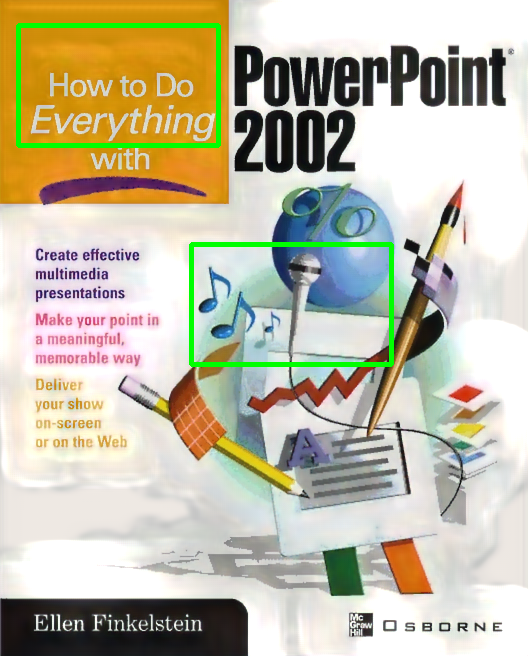}		
		\caption*{\centering{RMFNet~\cite{rmfnet_csvt2021}~\linebreak{(21.54/0.8567/10.69)}}}
	\end{subfigure}
	\begin{subfigure}{0.16\linewidth}
		\centering		
		\includegraphics[width=\linewidth]{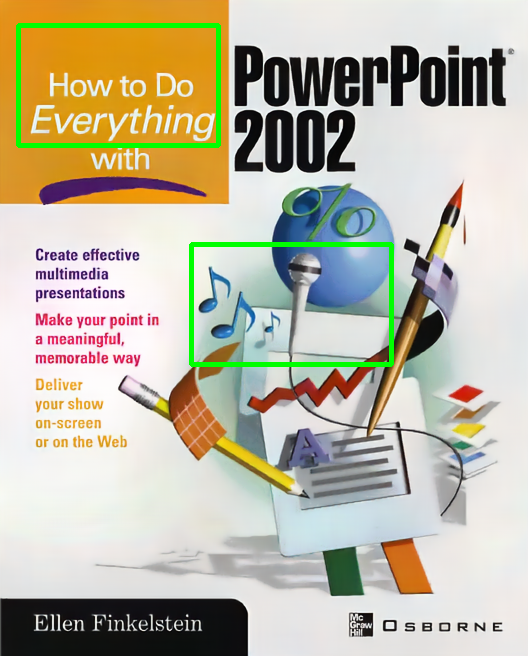}		
		\caption*{\centering{BRNet~\linebreak{(\textbf{23.41/0.9176/11.87})}}}
	\end{subfigure}		
	\caption{Visual comparison on restoring image from 2-bit to 8-bit on Set14 dataset. Please magnify the result for a better view.}
	\label{fig:vis-set14}
\end{figure*}

Figure \ref{fig:vis-set14} shows the visual comparison on restoring 2-bit image to 8-bit on Set14 dataset. In the figure, BRNet outputs fewer artifacts around the logo than BE-CALF, while LBDEN and RMFNet suffer from the severe false contours. All the competitive methods exhibit obvious banding effect on the blue circle within the green rectangle, while our BRNet has perfect color transition with accurate textures.

\begin{table*}[t]
	\centering
	\small
	
	\begin{tabular}{|c|ccc|c|}
		\hline
		\textbf{Bit-Depth}& \textbf{BE-CALF~\cite{be_calf_tip2019}}& \textbf{LBDEN~\cite{lbden_tcsvt2021}}& \textbf{RMFNet~\cite{rmfnet_csvt2021}} &\textbf{BRNet}\\	
		\hline
		\textbf{1}&	19.04 / 0.6603 / 15.68&	\textbf{19.67} / 0.6565 / \textbf{12.37}&	18.98 / 0.6598 / 16.06 &	18.86 / \textbf{0.6719} / 18.74\\
		\textbf{2}&	25.61 / 0.8162 / 6.93&	25.96 / 0.8191 / 6.47&	25.87 / 0.8230 / 6.18 &	\textbf{27.05 / 0.8470 / 5.06}\\
		\textbf{3}&	32.07 / 0.9104 / 3.00&	32.28 / 0.9155 / 2.46&	32.11 / 0.9204 / 2.23 &	\textbf{34.07 / 0.9460 / 1.20}\\
		\textbf{4}&	36.59 / 0.9501 / 2.08&	36.86 / 0.9625 / 1.33&	36.35 / 0.9572 / 1.51 &	\textbf{38.80 / 0.9740 / 0.82}\\
		\textbf{5}&	40.27 / 0.9631 / 2.29&	41.18 / 0.9840 / 1.42&	40.05 / 0.9766 / 1.68 &	\textbf{43.22 / 0.9749 / 0.64}\\
		\textbf{6}&	43.81 / 0.9639 / 2.38&	44.50 / 0.9841 / 1.60&	43.48 / 0.9827 / 1.69 &	\textbf{47.93 / 0.9965 / 0.30}\\
		\textbf{7}&	46.37 / 0.9649 / 2.49&	47.57 / 0.9880 / 1.53&	45.63 / 0.9838 / 1.74 &	\textbf{52.83 / 0.9987 / 0.10}\\
		\hline
	\end{tabular}
	
	\caption{PSNR/SSIM/W-dis performance comparison with state-of-the-art methods on restoring images with different bit depth on Set5 dataset. The \textbf{bold} value means the best performance.}
	\label{tab:set5}
\end{table*}

\begin{table*}[t]
	\centering
	\small
	
	\begin{tabular}{|c|ccc|c|}
		\hline
		\textbf{Bit-Depth}& \textbf{BE-CALF~\cite{be_calf_tip2019}}& \textbf{LBDEN~\cite{lbden_tcsvt2021}}& \textbf{RMFNet~\cite{rmfnet_csvt2021}}& \textbf{BRNet} \\	
		\hline
		\textbf{1}&	17.96 / 0.6174 / 18.79&	18.34 / 0.6169 / 14.69&	18.40 / 0.6333 / 16.49&	\textbf{18.60 / 0.6487 / 15.92}\\
		\textbf{2}&	23.83 / 0.7862 / 10.33&	24.79 / 0.8084 / 5.93&	24.73 / 0.8022 / 5.56&	\textbf{26.21 / 0.8348 / 4.50}\\
		\textbf{3}&	29.40 / 0.8966 / 5.84&	30.56 / 0.9153 / 3.09&	29.83 / 0.9078 / 3.22&	\textbf{32.79 / 0.9362 / 1.71}\\
		\textbf{4}&	33.94 / 0.9505 / 3.85&	34.47 / 0.9608 / 2.30&	33.68 / 0.9544 / 2.58&	\textbf{37.64 / 0.9739 / 1.06}\\
		\textbf{5}&	37.88 / 0.9688 / 3.07&	36.56 / 0.9761 / 2.75&	36.32 / 0.9698 / 2.93&	\textbf{42.11 / 0.9837 / 0.80}\\
		\textbf{6}&	41.06 / 0.9740 / 3.09&	37.69 / 0.9792 / 3.31&	38.23 / 0.9760 / 3.22&	\textbf{47.21 / 0.9965 / 0.43}\\
		\textbf{7}&	43.49 / 0.9753 / 3.18&	37.94 / 0.9807 / 3.70&	38.77 / 0.9777 / 3.56&	\textbf{52.19 / 0.9987 / 0.17}\\
		\hline
	\end{tabular}
	
	\caption{PSNR/SSIM/W-dis performance comparison with state-of-the-art methods on restoring images with different bit depth on Set14 dataset. The \textbf{bold} value means the best performance.}
	\label{tab:set14}
\end{table*}

\begin{table*}[t]
	\centering
	\small
	
	\begin{tabular}{|c|ccc|c|}
		\hline
		\textbf{Bit-Depth}& \textbf{BE-CALF~\cite{be_calf_tip2019}}& \textbf{LBDEN~\cite{lbden_tcsvt2021}}& \textbf{RMFNet~\cite{rmfnet_csvt2021}}& \textbf{BRNet} \\	
		\hline
		\textbf{1}&19.41 / 0.6423 / 15.12&	19.17 / 0.6285 / 15.28&	18.96 / 0.6330 / 15.65&	\textbf{19.91 / 0.6724 / 13.82}\\
		\textbf{2}&	26.41 / 0.8518 / 5.57&	26.49 / 0.8529 / 5.21&	26.40 / 0.8496 / 5.05&	\textbf{28.27 / 0.8811 / 3.57}\\
		\textbf{3}&	32.16 / 0.9347 / 3.22&	32.88 / 0.9425 / 1.92&	32.68 / 0.9390 / 1.88&	\textbf{34.62 / 0.9553 / 1.16}\\
		\textbf{4}&	37.38 / 0.9715 / 1.88&	38.42 / 0.9772 / 0.99&	38.25 / 0.9755 / 0.94&	\textbf{39.72 / 0.9807 / 0.67}\\
		\textbf{5}&	42.47 / 0.9897 / 1.09&	43.56 / 0.9922 / 0.58&	43.33 / 0.9911 / 0.56&	\textbf{44.35 / 0.9932 / 0.42}\\
		\textbf{6}&	46.90 / 0.9960 / 0.70&	47.83 / 0.9969 / 0.39&	47.47 / 0.9965 / 0.37&	\textbf{48.42 / 0.9972 / 0.32}\\
		\textbf{7}&	50.16 / 0.9981 / 0.53&	50.92 / 0.9985 / 0.26&	50.24 / 0.9983 / 0.28&	\textbf{53.30 / 0.9989 / 0.12}\\		
		\hline
	\end{tabular}
	
	\caption{PSNR/SSIM/W-dis performance comparison with state-of-the-art methods on restoring images with different bit depth on B100 dataset. The \textbf{bold} value means the best performance.}
	\label{tab:b100}
\end{table*}

\section{Conclusion}
In this paper, we proposed a novel network for bit-depth expansion (BDE) for any bit-depth degradation, termed as BRNet. Different from directly finding suitable values for high bit-depth (HBD) images from low bit-depth (LBD) instance, we investigated the BDE in an optimization perspective and learned a weighting map to replenish LBD image to HBD within a rational range. We designed an UNet-style network for missing bits restoration, which is termed as BRNet. Specially, proximal gradient descent algorithm and Runge-Kutta method were utilized for designing the block. Progressive training was considered in the training phase to improve the capacity. Experimental results show our BRNet achieves better subjective and objective performances than state-of-the-arts on all degradation situations. When the number of missing bits is larger, BRNet is much superior to the other works with satisfactory visual performance.

In furture, we try to focus on the texture restoration in BDE. Suffering from the severe information loss, the restored images of the network may loss some texture details from the ground truth when the number of missing bits is too large, as observed in the wings of parrot and the sky over sea in \cref{fig:wdis}. How to apply the structure information as a prior for effective restoration is an open issue for BDE.

{\small
	\bibliographystyle{ieee_fullname}
	\bibliography{egbib}
}

\vfill

\end{document}